\documentclass[a4paper,fleqn]{cas-dc}

\usepackage[numbers]{natbib}
\usepackage{makecell}
\usepackage{pdflscape,array,booktabs}
\usepackage{enumitem}
\usepackage{outlines}
\usepackage{tabularx}
\usepackage{longtable}
\usepackage{mdframed}
\usepackage{soul}
\usepackage{xurl}
\usepackage{subfigure}
\usepackage{multirow}
\usepackage[parfill]{parskip}
\usepackage{caption}

\usepackage[textsize=scriptsize,backgroundcolor=yellow!40]{todonotes}

\def\tsc#1{\csdef{#1}{\textsc{\lowercase{#1}}\xspace}}
\tsc{WGM}
\tsc{QE}
\tsc{EP}
\tsc{PMS}
\tsc{BEC}
\tsc{DE}

\begin{document}
\let\WriteBookmarks\relax
\def\floatpagepagefraction{1}
\def\textpagefraction{.001}
\shorttitle{Architectural Patterns for the Design of Federated Learning Systems}
\shortauthors{SK Lo et~al.}

\title [mode = title]{Architectural Patterns for the Design of Federated Learning Systems}

\author[1,2]{Sin Kit Lo}[type=editor,
                        auid=000,bioid=1,
                        orcid=0000-0002-9156-3225]
\cormark[1]

\ead{Kit.Lo@data61.csiro.au}

\author[1,2]{Qinghua Lu}[type=editor,
                        auid=000,bioid=1,
                        orcid=0000-0002-7783-5183]

\ead{Qinghua.Lu@data61.csiro.au}

\author[1,2]{Liming Zhu}[type=editor,
                        auid=000,bioid=1,]

\ead{liming.zhu@data61.csiro.au}

\author[2]{Hye-Young Paik}[type=editor,
                        auid=000,bioid=1,
                        orcid=0000-0003-4425-7388]

\ead{h.paik@unsw.edu.au}

\author[1,2]{Xiwei Xu}[type=editor,
                        auid=000,bioid=1,
                        orcid=0000-0002-2273-1862]

\ead{Xiwei.Xu@data61.csiro.au}

\author[1]{Chen Wang}[type=editor,
                        auid=000,bioid=1,
                        orcid=0000-0002-3119-4763]

\ead{Chen.Wang@data61.csiro.au}

\address[1]{Data61, CSIRO, Australia}

\address[2]{University of New South Wales, Australia}

\begin{abstract}
Federated learning has received fast-growing interests from academia and industry to tackle the challenges of data hungriness and privacy in machine learning. A federated learning system can be viewed as a large-scale distributed system with different components and stakeholders as numerous client devices participate in federated learning. Designing a federated learning system requires software system design thinking apart from the machine learning knowledge. Although much effort has been put into federated learning from the machine learning technique aspects, the software architecture design concerns in building federated learning systems have been largely ignored. Therefore, in this paper, we present a collection of architectural patterns to deal with the design challenges of federated learning systems. Architectural patterns present reusable solutions to a commonly occurring problem within a given context during software architecture design. The presented patterns are based on the results of a systematic literature review and include three client management patterns, four model management patterns, three model training patterns, and four model aggregation patterns. The patterns are associated to the particular state transitions in a federated learning model lifecycle, serving as a guidance for effective use of the patterns in the design of federated learning systems.

\end{abstract}



\maketitle

\section{Introduction}
\label{S:1}

\subsection*{Federated Learning Overview}

The ever-growing use of big data systems, industrial-scale IoT platforms, and smart devices contribute to the exponential growth in data dimensions~\cite{s19204354}. This exponential growth of data has accelerated the adoption of machine learning in many areas, such as natural language processing and computer vision. However, many machine learning systems suffer from insufficient training data. The reason is mainly due to the increasing concerns on data privacy, e.g., restrictions on data sharing with external systems for machine learning purposes. For instance, the General Data Protection Regulation (GDPR)~\cite{(gdpr)_2019} stipulate a range of data protection measures, and data privacy is now one of the most important ethical principles expected of machine learning systems \cite{jobin2019global}. Furthermore, raw data collected are unable to be used directly for model training for most circumstances. The raw data needs to be studied and pre-processed before being used for model training and data sharing restrictions increase the difficulty to obtain training data. Moreover, concept drift~\cite{9084352} also occurs when new data is constantly collected, replacing the outdated data. This makes the model trained on previous data degrades at a much faster rate. Hence, a new technique that can swiftly produce models that adapt to the concept drift when different data is discovered in clients is essential.      

\begin{figure}[tbh!]
\centering\includegraphics[width=0.9\linewidth]{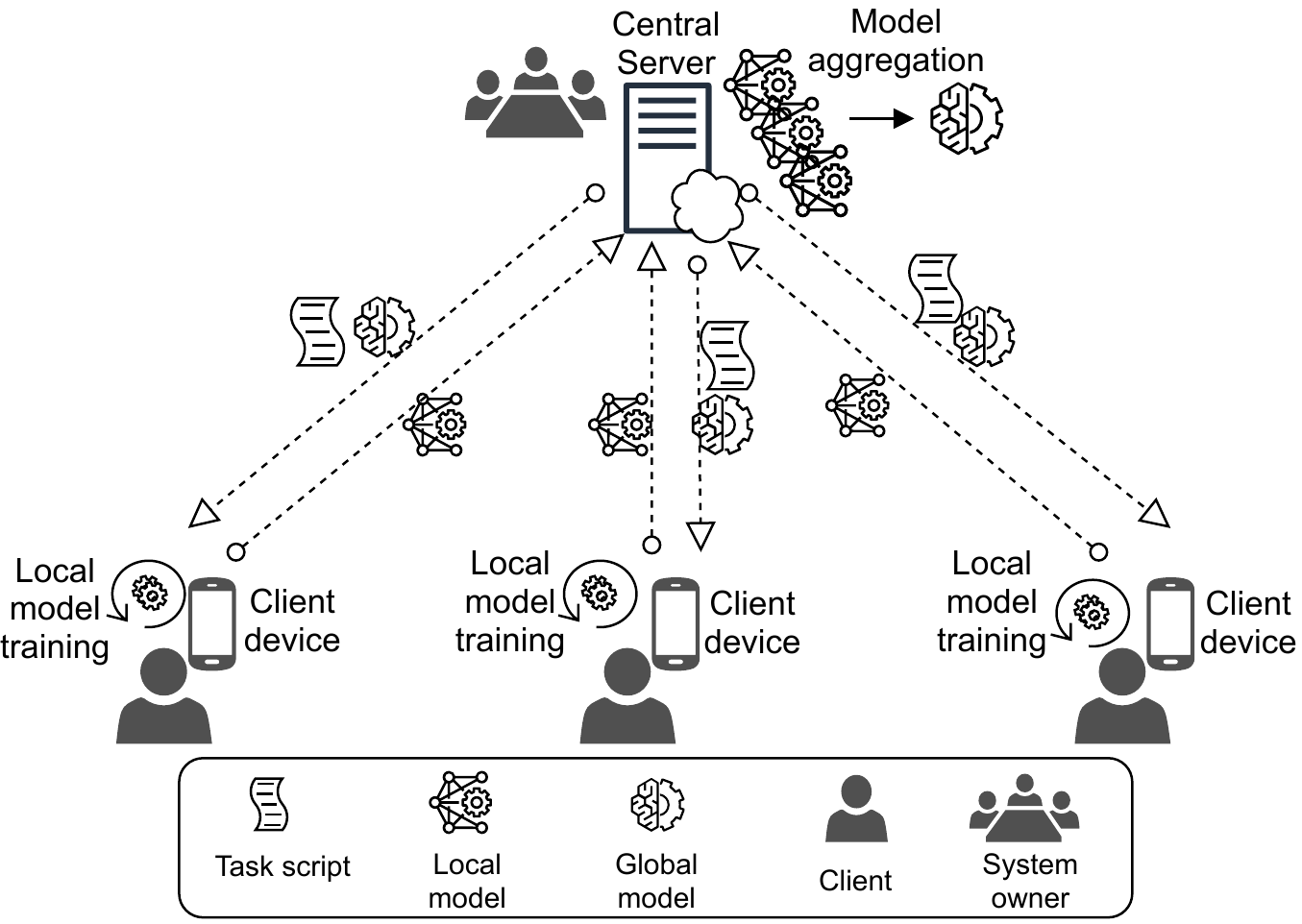}
\caption{Federated Learning Overview.}
\label{Fig:FLOps}
\end{figure}

To effectively address the lack of training data limitations, concept drift, and the data-sharing restriction while still enabling effective data inferences by the data-hungry machine learning models, Google introduced the concept of federated learning~\cite{mcmahan2017communicationefficient} in 2016. Fig.~\ref{Fig:FLOps} illustrates the overview of federated learning. There are three stakeholders in a federated learning system: (1) learning coordinator (i.e., system owner), (2) contributor client - data contributor \& local model trainer, and (3) user client - model user. Note that a contributor client can also be a user client. There are two types of system nodes (i.e., hardware components): (1) central server, (2) client device.

\subsection*{A Motivation Example}
Imagine we use federated learning to train the next-word prediction model in a mobile phone keyboard application. The learning coordinator is the provider of the keyboard application, while contributor clients are the mobile phone users. The user clients will be the new or existing mobile phone users of the keyboard application. The differences in ownership, geolocation, and usage pattern cause the local data to possess non-IID\footnote{Non-Identically and Independently Distribution: Highly-skewed and personalised data distribution that vary heavily between different clients and affects the model performance and generalisation~\cite{8889996}.} characteristics, which is a design challenge of federated learning systems. The federated learning process starts when a training task is created by the learning coordinator. For instance, the keyboard application provider produces and embeds an initial global model into the keyboard applications. The initial global model (includes task scripts \& training hyperparameters) is broadcast to the participating client devices. After receiving the initial model, the model training is performed locally across the client devices, without the centralised collection of raw client data. Here, the smartphones that have the keyboard application installed receive the model to be trained. The client devices typically run on different operating systems and have diverse communication and computation resources, which is defined as the system heterogeneity challenges.

Each training round takes one step of gradient descent on the current model using each client's local data. In this case, the smartphones optimise the model using the keyboard typing data. After each round, the model update is submitted by each participating client device to the central server. The central server collects all the model updates and performs model aggregation to form a new version of the global model. The new global model is re-distributed to the client devices for the next training round. This entire process iterates until the global model converges. As a result, communication and computation efficiency are crucial as many local computation and communication rounds are required for the global model to converge. Moreover, due to the limited resources available on each device, the device owners might be reluctant to participate in the federated learning process. Therefore, client motivatability becomes a design challenge. Furthermore, the central server communicates with multiple devices for the model exchange which makes it vulnerable to the single-point-of-failure. The trustworthiness of the central server and the possibility of adversarial nodes participating in the training process also creates system reliability and security challenges. After the completion of training, the learning coordinator stores the converged model and deploys it to the user clients. For instance, the keyboard application provider stores the converged model and embeds it to the latest version of the application for existing or new application users. Here, the different versions of the local models associated with the global model created need to be maintained.

\subsection*{Design Challenges}

Compared to centralised machine learning, federated learning is more advantageous from the data privacy perspective and dealing with the lack of training data. However, a federated learning system, as a large-scale distributed system, presents more architectural design challenges~\cite{10.1145/3450288}, especially when dealing with the interactions between the central server and client devices and managing trade-offs of software quality attributes. The main design challenges are summarised as follows. 
\begin{itemize}[leftmargin=*]
\item 
Global models might have low accuracy, and lack generality when client devices generate non-IID data. Centralising and randomising the data is the approach adopted by conventional machine learning to deal with data heterogeneity but the inherent privacy-preserving nature of federated learning render such techniques inappropriate.
\item  To generate high-quality global models that are adaptive to concept drift, multiple rounds of communication are required to exchange local model updates, which could incur high communication costs.
\item  Client devices have limited resources to perform the multiple rounds of model training and communications required by the system, which may affect the model quality. 
\item As numerous client devices participate in federated learning, it is challenging to coordinate the learning process and ensure model provenance, system reliability and security.
\end{itemize}

How to select appropriate designs to fulfill different software quality requirements and design constraints is non-trivial for such a complex distributed system. Although much effort has been put into federated learning from the machine learning techniques side, there is still a gap on the architectural design considerations of the federated learning systems. A systematic guidance on architecture design of federated learning systems is required to better leverage the existing solutions and promote federated learning to enterprise-level adoption.

\begin{figure*}
\centering
\includegraphics[width=0.75\linewidth]{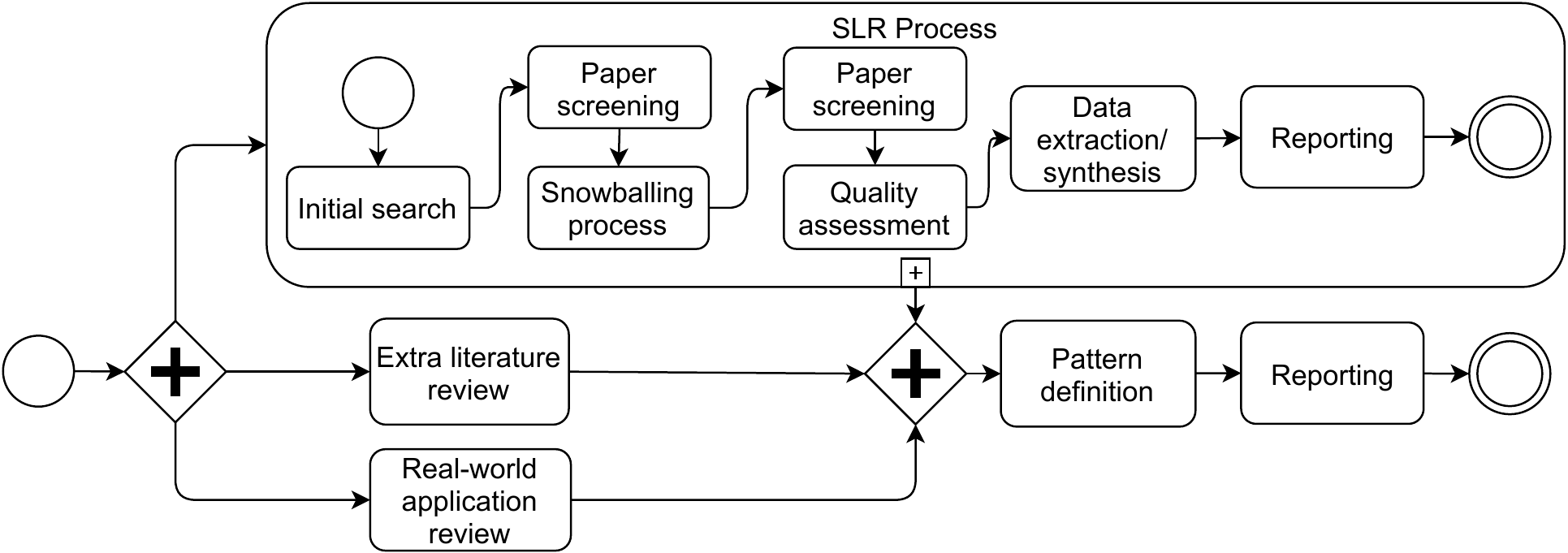}
\caption{Pattern Collection Process.}
\label{Fig:CollectionProtocol}
\end{figure*}

\subsection*{Research Contribution}

In this paper, we present a collection of patterns for the design of federated learning systems. In software engineering, an architectural pattern is a reusable solution to a problem that occurs commonly within a given context in software design~\cite{Beck1987}. Our pattern collection includes three client management patterns, four model management patterns, three model training patterns, and four model aggregation patterns. We define the lifecycle of a model in a federated learning system and associate each identified pattern to a particular state transition in the lifecycle.

The main contribution of this paper includes:

\begin{itemize}[leftmargin=*]
\item The collection of architectural patterns provides a design solution pool for practitioners to select from for real-world federated learning system development.
\item The federated learning model lifecycle with architectural pattern annotations, which serves as a systematic guide for practitioners during the design and development of a federated learning system.
 
\end{itemize}

\subsection*{Paper Structure}

The remainder of the study is organised as follows. Section~\ref{S:methodology} introduces the research methodology. Section~\ref{S:patterns} provides an overview of the patterns in the federated learning lifecycle, followed by the detailed discussions on each pattern. Section~\ref{S:discussion} summarises and discusses some repeating challenges of federated learning systems and Section~\ref{S:related} presents the related work. Finally, Section~\ref{S:conclusion} concludes the paper.

\begin{table*}[!h]

\scriptsize
\linespread{1.3}
\centering
\caption{Sources of Patterns.}
\begin{tabular}{l l c c c}
\toprule
\textbf{\footnotesize{Category}} & \textbf{\makecell[c]{\footnotesize{Pattern}}} & \textbf{\makecell[l]{\footnotesize{SLR papers}}} & \textbf{\makecell[l]{\footnotesize{ML/FL papers}}} & \textbf{\makecell[l]{\footnotesize{Real-world applications}}}\\
\midrule

\multirow{3}{0.2\columnwidth}{Client management patterns} & Pattern 1: Client registry & 0 & 2 & 3\\
& Pattern 2: Client selector & 4 & 2 & 1\\
& Pattern 3: Client cluster & 2 & 2 & 2\\
\cmidrule(l){1-5}

\multirow{4}{0.2\columnwidth}{Model management patterns} & Pattern 4: Message compressor & 8 & 6 & 0\\
 & Pattern 5: Model co-versioning registry & 0 & 0 & 4\\
 & Pattern 6: Model replacement trigger & 0 & 1 & 3\\
 & Pattern 7: Deployment selector & 0 & 0 & 3\\
\cmidrule(l){1-5}
\multirow{3}{0.2\columnwidth}{Model training patterns} & Pattern 8: Multi-task model trainer & 2 & 1 & 3\\
& Pattern 9: Heterogeneous data handler & 1 & 2 & 0\\
& Pattern 10: Incentive registry & 18 & 1 & 0\\
\cmidrule(l){1-5}
\multirow{4}{0.2\columnwidth}{Model aggregation patterns} & Pattern 11: Asynchronous aggregator & 4 & 1 & 0\\
& Pattern 12: Decentralised aggregator & 5 & 2 & 0\\
& Pattern 13: Hierarchical aggregator & 4 & 2 & 0\\
& Pattern 14: Secure aggregator & 31 & 0 & 3\\
\bottomrule
\label{tab:mapping}
\end{tabular}

\end{table*}

\section{Methodology}
\label{S:methodology}

Fig.~\ref{Fig:CollectionProtocol} illustrates the federated learning pattern extraction and collection process. Firstly, the patterns are collected based on the results of our previous systematic literature review (SLR) on federated learning~\cite{10.1145/3450288}. SLR and situational method engineering (SME)~\cite{BRINKKEMPER1996275} are some of the renowned systematic methodologies for  derivation of pattern languages. For instance, several pattern derivations on cloud migration and software architecture have used SLR (e.g., Zdun et al.~\cite{zdun2007systematic}, Aakash Ahmad et al.~\cite{ahmad2014pattern}, and Jamshidi et al.~\cite{jamshidi2017pattern}). Moreover, Balalaie et al.~\cite{balalaie2018microservices} have derived the pattern languages in the context of cloud-native and microservices using situational method engineering. 

For this work, we have adopted the SLR method as the currently available materials and research works on federated learning are still highly academic-based. Secondly, we intend to propose design patterns for software architectural design aspects of building federated learning systems rather than for their development/engineering processes. This is because, during the SLR work, we have identified many architectural design challenges and lack of systematic design approaches to federated learning. Furthermore, while SME has the benefit of offering a systematic methodology for selecting appropriate method components from a repository of reusable method components, it is more suitable for pattern extraction of an information system development (ISD) process~\cite{mirbel2006situational}.

The SLR was performed according to Kitchenham’s SLR guideline~\cite{Kitchenham07guidelinesfor}, and the number of final studied papers is 231. During the SLR, we developed a comprehensive mapping between federated learning challenges and approaches. Additionally, we reviewed 22 machine learning and federated learning papers published after the SLR search cut-off date (31st Jan 2020) and 22 real-world applications. The additional literature review on machine learning and federated learning, and the review of the real-world applications are conducted based on our past real-world project implementation experience. Table~\ref{tab:mapping} shows a mapping between each pattern with its respective number of source papers and real-world applications. Twelve patterns were initially collected from SLR or additional literature review, whereas the remaining two patterns were identified through real-world applications.

We discussed the proposed patterns according to the pattern form presented in \cite{10.5555/273448.273487}. The form comprehensively describes the patterns by discussing the \textbf{Context}, \textbf{Problem}, \textbf{Forces}, \textbf{Solution}, \textbf{Consequences}, \textbf{Related patterns}, and \textbf{Known-uses} of the pattern. 

The \textbf{Context} is the description of the situation where a problem occurs, in which the solution proposed is applicable, or the problem is solvable by the pattern. \textbf{Problem} comprehensively elicits the challenges and limitations that occur under the defined context. \textbf{Forces} describe the reasons and causes for a specific design or pattern decision to be made to resolve the problem. 
\textbf{Solution} describes how the problem and the conflict of forces can be resolved by a specific pattern. \textbf{Consequences} reason about the impact of applying a solution, specifically on the contradictions among the benefits, costs, drawbacks, tradeoffs, and liabilities of the solution. \textbf{Related patterns} record the other patterns from this paper that are related to the current pattern. \textbf{Known-uses} refer to empirical evidence that the solution has been used in the real world.

\section{Federated Learning Patterns}
\label{S:patterns}

\begin{figure*}
\centering\includegraphics[width=0.8\linewidth]{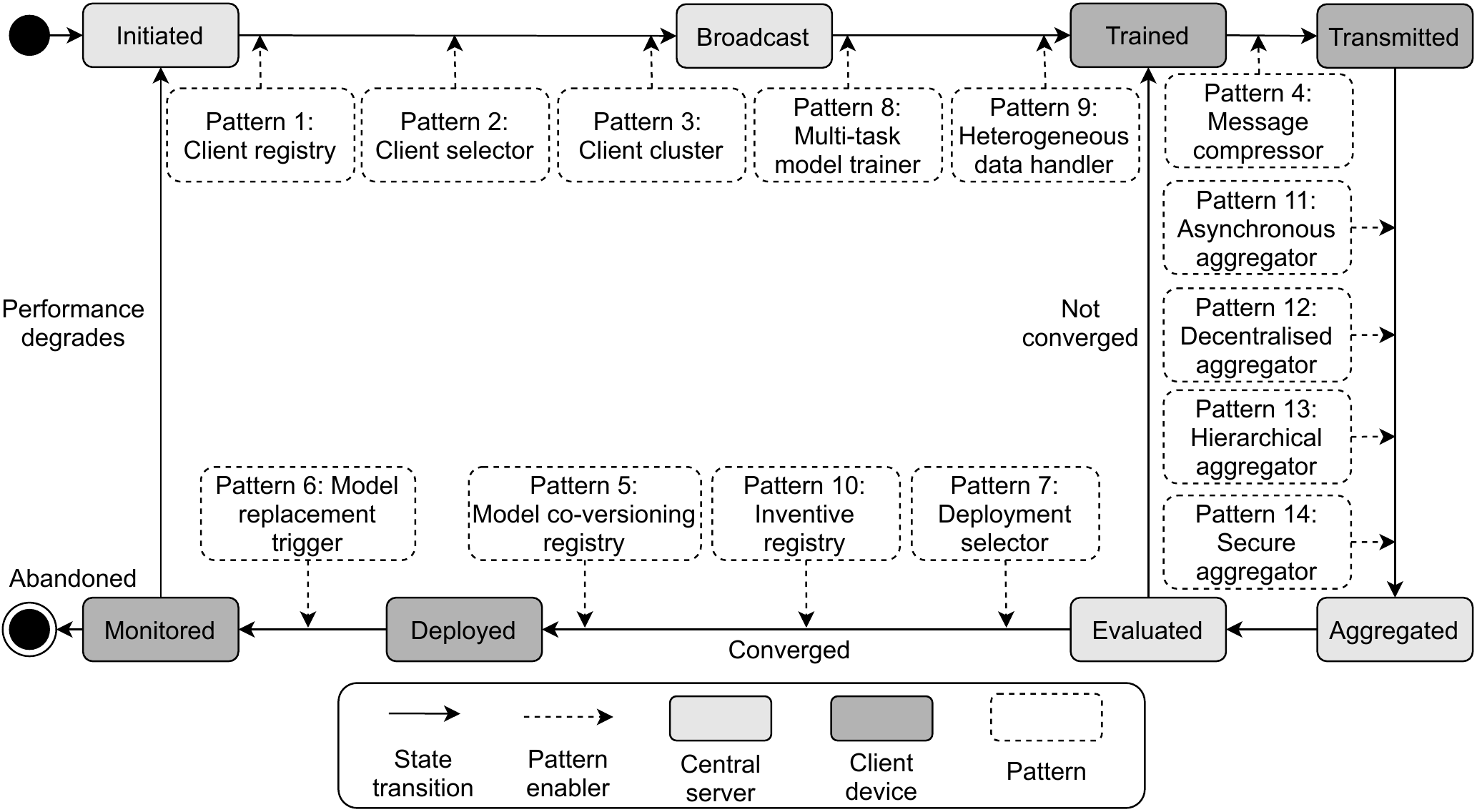}
\caption{A Model's Lifecycle in Federated Learning.}
\label{Fig:FLLifeCycle}
\end{figure*}

Fig.~\ref{Fig:FLLifeCycle} illustrates the lifecycle of a model in a federated learning system. The lifecycle covers the deployment of the completely trained global model to the client devices (i.e., model users) for data inference. The deployment process involves the communication between the central server and the client devices. We categorise the federated learning patterns as shown in Table \ref{tab:overview} to provide an overview. There are four main groups: (1) client management patterns, (2) model management patterns, (3) model training patterns, and (4) model aggregation patterns.

\begin{table*}[tbp]
\scriptsize
\centering
\caption{Overview of architectural patterns for federated learning}
\label{tab:overview}
\begin{tabular}{p{0.3\columnwidth}p{0.4\columnwidth}p{1.2\columnwidth}}
\toprule

\textbf{\makecell[l]{\footnotesize{Category}}} &
\textbf{\makecell[c]{\footnotesize{Name}}} &
\textbf{\makecell[c]{\footnotesize{Summary}}}\\
\midrule

\multirow{5}{0.3\columnwidth}{\textbf{Client management patterns}} & \multirow{1}{0.32\columnwidth}{Client registry} & Maintains the information of all the participating client devices for client management.\\
\cmidrule(l){2-3}

& \multirow{1}{0.32\columnwidth}{Client selector} & Actively selects the client devices for a certain round of training according to the predefined criteria to increase model performance and system efficiency.\\
\cmidrule(l){2-3}

& \multirow{1}{0.32\columnwidth}{Client cluster} & Groups the client devices (i.e., model trainers) based on their similarity of certain characteristics (e.g., available resources, data distribution, features, geolocation) to increase the model performance and training efficiency.\\

\cmidrule(l){1-3}
\cmidrule(l){1-3}
\multirow{8}{0.3\columnwidth}{\textbf{Model management patterns}} & \multirow{2}{0.32\columnwidth}{Message compressor} & Compresses and reduces the message data size before every round of model exchange to increase the communication efficiency.\\

\cmidrule(l){2-3}
& \multirow{1}{0.4\columnwidth}{Model co-versioning registry} & Stores and aligns the local models from each client with the corresponding global model versions for model provenance and model performance tracking.\\
\cmidrule(l){2-3}
& \multirow{1}{0.4\columnwidth}{Model replacement trigger} & Triggers model replacement when the degradation in model performance is detected.\\
\cmidrule(l){2-3}
& \multirow{1}{0.32\columnwidth}{Deployment selector} & Selects and matches the converged global models to suitable client devices to maximise the global models' performance for different applications and tasks.\\

\cmidrule(l){1-3}
\cmidrule(l){1-3}
\multirow{8}{0.3\columnwidth}{\textbf{Model training patterns}} & \multirow{1}{0.4\columnwidth}{Multi-task model trainer} & Utilises data from separate but related models on local client devices to improve learning efficiency and model performance.\\
\cmidrule(l){2-3}
& \multirow{1}{0.4\columnwidth}{Heterogeneous data handler} & Solves the non-IID and skewed data distribution issues through data volume and data class addition while maintaining the local data privacy.\\
\cmidrule(l){2-3}
& \multirow{1}{0.32\columnwidth}{Incentive registry} & Measures and records the performance and contributions of each client and provides incentives to motivate clients' participation. \\

\cmidrule(l){1-3}
\multirow{8}{0.3\columnwidth}{\textbf{Model aggregation patterns}} & \multirow{1}{0.4\columnwidth}{Asynchronous aggregator} & Performs aggregation asynchronously whenever a model update arrives without waiting for all the model updates every round to reduce aggregation latency.\\
\cmidrule(l){2-3}
& \multirow{1}{0.4\columnwidth}{Decentralised aggregator} & Removes the central server from the system and decentralizes its role to prevent single-point-of-failure and increase system reliability.\\
\cmidrule(l){2-3}
& \multirow{1}{0.32\columnwidth}{Hierarchical aggregator} & Adds an edge layer to perform partial aggregation of local models from closely-related client devices to improve model quality and system efficiency. \\
\cmidrule(l){2-3}
& \multirow{1}{0.32\columnwidth}{Secure aggregator} 
& The adoption of secure multiparty computation (MPC) protocols that manages the model exchange and aggregation security to protect model security. \\
\bottomrule
\end{tabular}
\end{table*}

\subsection{\textbf{Client Management Patterns}}\label{ClientManagementPatterns}
Client management patterns describe the patterns that manage the client devices' information and their interaction with the central server. A \textit{client registry} manages the information of all the participating client devices. \textit{Client selector} selects client devices for a specific training task, while \textit{client cluster} increases the model performance and training efficiency through grouping client devices based on the similarity of certain characteristics (e.g., available resources, data distribution).

\subsubsection{\textbf{Pattern 1: Client Registry}}\label{ClientRegistry}

\begin{figure}[h]
\centering\includegraphics[width=0.7\linewidth]{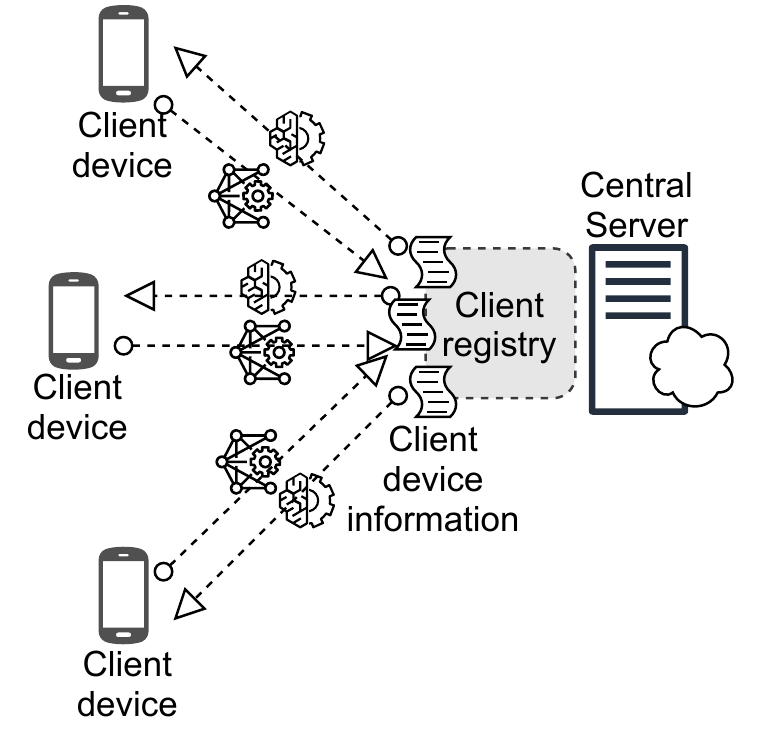}

\caption{Client Registry.}
\label{Fig:ClientRegistry}
\end{figure}

\textbf{Summary:} A client registry maintains the information of all the participating client devices for client management. According to Fig.~\ref{Fig:ClientRegistry}, the client registry is maintained in the central server. The central server sends the request for information to the client devices. The client devices then send the requested information together with the first local model updates.

\textbf{Context:} Client management is centralised, and global and local models are exchanged between the central server the massive number of distributed client devices with dynamic connectivity and diverse resources.

\textbf{Problem:} It is challenging for a federated learning system to track any dishonest, failed, or dropout node. This is crucial to secure the central server and client devices from adversarial threats. Moreover, to effectively align the model training process of each client device for each aggregation round, a record of the connection and training information of each client device that has interacted with the central server is required. 

\textbf{Forces:} The problem requires to balance the following forces:

\begin{itemize}[leftmargin=*]
    \item \textit{Updatability.} The ability to keep track of the participating devices is necessary to ensure the information recorded is up-to-date.

    \item \textit{Data privacy.} The records of client information expose the clients to data privacy issues. For instance, the device usage pattern of users may be inferred from the device connection up-time, device information, resources, etc.

\end{itemize}

\textbf{Solution:} A client registry records all the information of client devices that are connected to the system from the first time. The information includes device ID, connection up \& downtime, device resource information (computation, communication, power \& storage). The access to the client registry could be restricted according to the agreement between the central server and participating client devices.

\textbf{Consequences:}

Benefits:
\begin{itemize}[leftmargin=*]
    \item \textit{Maintainability.} The client registry enables the system to effectively manage the dynamically connecting and disconnecting clients. 
    
    \item \textit{Reliability.} The client registry provides status tracking for all the devices, which is essential for problematic node identification. 
    
\end{itemize}

Drawbacks:

\begin{itemize}[leftmargin=*]
    \item \textit{Data privacy.} The recording of the device information on the central server leads to client data privacy issues. 
    \item \textit{Cost.} The maintenance of client device information requires extra communication cost and storage cost, which further surges when the number of client devices increases.

\end{itemize}

\textbf{Related patterns:} \textit{Model Co-Versioning Registry, Client Selector, Client Cluster, Asynchronous Aggregator, Hierarchical Aggregator}

\textbf{Known uses:}
\begin{itemize}[leftmargin=*]
 \item \textit{IBM Federated Learning\footnote{\url{https://github.com/IBM/federated-learning-lib}}}: \textit{Party Stack} component manages the client parties of IBM federated learning framework, that contains sub-components such as protocol handler, connection, model, local training, and data handler for client devices registration and management.
 
 \item doc.ai\footnote{\url{https://doc.ai/}}: Client registry is designed for medical research applications to ensure that updates received apply to a current version of the global model, and not a deprecated global model.
 \item \textit{SIEMENS Mindsphere Asset Manager \footnote{\url{https://documentation.mindsphere.io/resources/html/asset-manager/en-US/index.html}}}:To support the collaboration of federated learning clients in industrial IoT environment,~\textit{Industrial Metadata Management} is introduced as a device metadata and asset data manager.

\end{itemize}

\subsubsection{\textbf{Pattern 2: Client Selector}}\label{ClientSelector}

\begin{figure}[h]
\centering\includegraphics[width=0.7\linewidth]{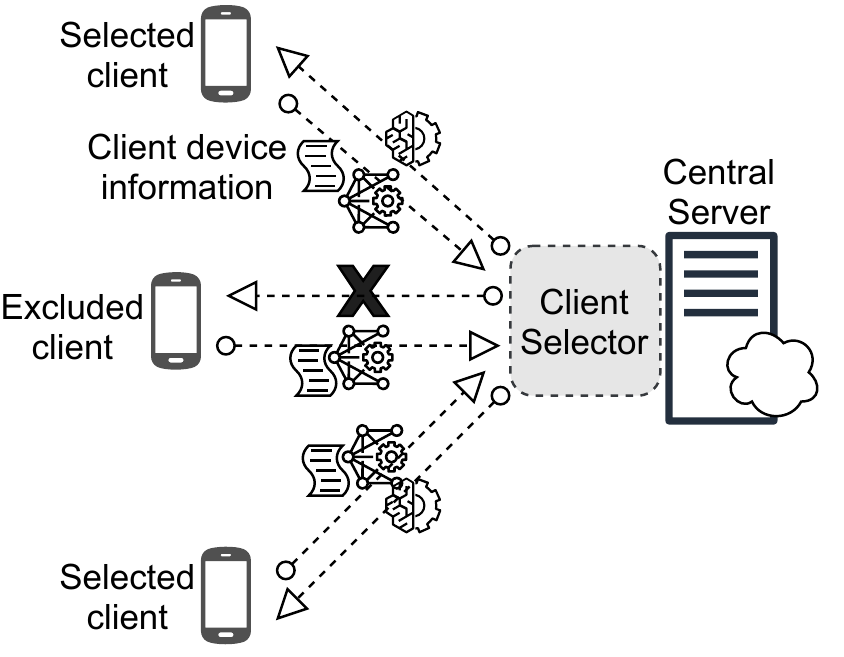}
\caption{Client Selector.}
\label{Fig:ClientSelector}
\end{figure}

\textbf{Summary:} A client selector actively selects the client devices for a certain round of training according to the predefined criteria to increase model performance and system efficiency. As shown in Fig.~\ref{Fig:ClientSelector}, the central server assesses the performance of each client according to the information received. Based on the assessment results, the second client is excluded from receiving the global model.

\textbf{Context:} Multiple rounds of model exchanges are performed and communication cost becomes a bottleneck. Furthermore, multiple iterations of aggregations are performed and consume high computation resources.

\textbf{Problem:} The central server is burdensome to accommodate the communication with massive number of widely-distributed client devices every round.

\textbf{Forces:} The problem requires the following forces to be balanced:
\begin{itemize}[leftmargin=*]
\item \textit{Latency.} Client devices have system heterogeneity (difference in computation, communication, \& energy resources) that affect the local model training and global model aggregation time. 

\item \textit{Model quality.} Local data are statistically heterogeneous (different data distribution/quality) which produce local models that overfit the local data.

\end{itemize}

\textbf{Solution:} Selecting client devices with predefined criteria can optimise the formation of the global model. The client selector on the central server performs client selection every round to include the best fitting client devices for global model aggregation. The selection criteria can be configured as follows:
\begin{itemize}[leftmargin=*]
\item Resource-based: The central server assesses the resources available on each client devices every training round and selects the client devices with the satisfied resource status (e.g., WiFi connection, pending status, sleep time) 

\item Data-based: The central server examines the information of the data collected by each client, specifically on the number of data classes, distribution of data sample volume per class, and data quality. Based on these assessments, the model training process  includes devices with high-quality data, higher data volume per class, and excludes the devices with low-quality data, or data that are highly heterogeneous in comparison with other devices.
\item Performance-based: Performance-based client selection can be conducted through local model performance assessment (e.g., performance of the latest local model or the historical records of local model performance). 

\end{itemize}

\textbf{Consequences:}

Benefits:
\begin{itemize}[leftmargin=*]
\item  \textit{Resource optimisation.} The client selection optimises the resource usage of the central server to compute and communicate with suitable client devices for each aggregation round, without wasting resources to aggregate the low-quality models.

\item \textit{System performance.} Selecting clients with sufficient power and network bandwidth greatly reduces the chances of clients dropping out and lowers the communication latency.

\item \textit{Model performance.} Selecting clients with the higher local model performance or lower data heterogeneity increases the global model quality.

\end{itemize}

Drawbacks:
\begin{itemize}[leftmargin=*]
\item \textit{Model generality.} The exclusion of models from certain client devices may lead to the missing of essential data features and the loss of the global model generality.

\item \textit{Data privacy.} The central server needs to acquire the system and resource information (processor's capacity, network availability, bandwidth, online status, etc.) every round to perform devices ranking and selection. Access to client devices' information creates data privacy issues.

\item \textit{Computation cost.} Extra resources are spent on transferring of the required information for selection decision-making.
\end{itemize}

\textbf{Related patterns:} \textit{Client Registry, Deployment Selector}

\textbf{Known uses:}
\begin{itemize}[leftmargin=*]
\item \textit{Google's FedAvg}: \textit{FedAvg} \cite{mcmahan2017communicationefficient} algorithm includes client selection that randomly selects a subset of clients for each round based on predefined environment conditions and device specification of the client devices.

\item In \textit{IBM's Helios} \cite{xu2019helios}, there is a training consumption profiling function that fully profiles the resource consumption for model training on client devices. Based on that profiling, a resource-aware soft-training scheme is designed to accelerate local model training on heterogeneous devices and prevent stragglers from delaying the collaborative learning process.

\item \textit{FedCS} suggested by OMRON SINIC X Corporation\footnote{\url{https://www.omron.com/sinicx/}} sets a certain deadline for clients to upload the model updates. 

\item \textit{Communication-Mitigated Federated Learning (CMFL)} \cite{8885054} excludes the irrelevant local updates by identifying the relevance of a client update by comparing its global tendency of model updating with all the other clients. 

\item \textit{CDW\_FedAvg}~\cite{9233457} takes the centroid distance between the positive and negative classes of each client dataset into account for aggregation.

\end{itemize}

\subsubsection{\textbf{Pattern 3: Client Cluster}}\label{ClientCluster}

\begin{figure}[h]
\centering\includegraphics[width=0.6\linewidth]{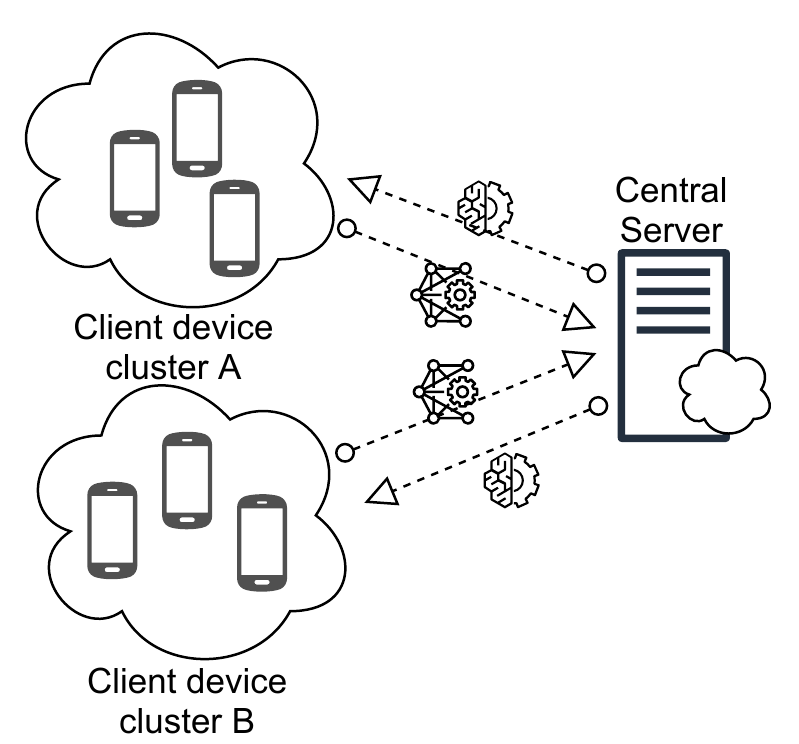}

\caption{Client Cluster.}
\label{Fig:ClientCluster}
\end{figure}

\textbf{Summary:} A client cluster groups the client devices (i.e., model trainers) based on their similarity of certain characteristics (e.g., available resources, data distribution, features, geolocation) to increase the model performance and training efficiency. In Fig.~\ref{Fig:ClientCluster}, the client devices are clustered into 2 groups by the central server, and the central server will broadcast the global model that is more related to the clusters accordingly.

\textbf{Context:} The system trains models over client devices which have diverse communication and computation resources, resulted in statistical and system heterogeneity challenges. 

\textbf{Problem:} Federated learning models generated under non-IID data properties are deemed to be less generalised. This is due to the lack of significantly representative data labels from the client devices. Furthermore, local models may drift significantly from each other.

\textbf{Forces:} The problem requires to balance the following forces:

\begin{itemize}[leftmargin=*]
    \item \textit{Computation cost and training time.} More computation costs and longer training time are required to overcome the non-IID issue of client devices. 

    \item \textit{Data privacy.} Data privacy contradicts with the Access to the entire or parts of the client's raw data is needed by the learning coordinator to resolve the non-IID issue which creates data privacy risks.

\end{itemize}

\textbf{Solution:} Client devices are clustered into different groups according to their properties (e.g., data distribution, features similarities, gradient loss). By creating clusters of clients with similar data patterns, the global model generated will have better performance for the non-IID-severe client network, without accessing the local data. 

\textbf{Consequences:}

Benefits:
\begin{itemize}[leftmargin=*]
    \item \textit{Model quality.} The global model created by client clusters can have a higher model performance for highly personalised prediction tasks.
    
    \item \textit{Convergence speed.} The consequent deployed global models can have faster convergence speed as the models of the same cluster can identify the gradient's minima much faster when the clients' data distribution and IIDness are similar.
\end{itemize}

Drawbacks:

\begin{itemize}[leftmargin=*]
    \item \textit{Computation cost.} The central server consumes extra computation cost and time for client clustering and relationship quantification.
    
    \item \textit{Data privacy.} The learning coordinator (i.e., central server) requires extra client device information (e.g., data distribution, feature similarities, gradient loss) to perform clustering. This exposes the client devices to the possible risk of private data leakage.
\end{itemize}

\textbf{Related patterns:} \textit{Client Registry, Client Selector, Deployment Selector}

\textbf{Known uses:}
\begin{itemize}[leftmargin=*]
 \item Iterative Federated Clustering Algorithm (\textit{IFCA})\footnote{\url{https://github.com/jichan3751/ifca}} is a \\framework introduced by UC Berkley and Google to cluster client devices based on the loss values of the client's gradient.
 
 \item Clustered Federated Learning (\textit{CFL})\footnote{\url{https://github.com/felisat/clustered-federated-learning}}
 uses a cosine simila\\rity-based clustering method that creates a bi-partitioning to group client devices with the same data generating distribution into the same cluster.
 
 \item \textit{TiFL} \cite{10.1145/3369583.3392686} is a tier-based federated learning system that adaptively groups client devices with similar training time per round to mitigate the heterogeneity problem without affecting the model accuracy. 
 
 \item Patient clustering in a federated learning system is implemented by Massachusetts General Hospital to improve efficiency in predicting mortality and hospital stay time \cite{HUANG2019103291}. 
\end{itemize}

\subsection{\textbf{Model Management Patterns}}\label{ModelManagementPatterns}
Model management patterns include patterns that handle mo\\-del transmission, deployment, and governance. A \textit{message compressor} reduces the transmitted message size. A \textit{model co-versioning registry} records all local model versions from each client and aligns them with their corresponding global model. A \textit{model replacement trigger} initiates a new model training task when the converged global model's performance degrades. A \textit{deployment selector} deploys the global model to the selected clients to improve the model quality for personalised tasks. 

\subsubsection{\textbf{Pattern 4: Message Compressor}}\label{MessageCompressor}

\begin{figure}[h]
\centering\includegraphics[width=0.85\linewidth]{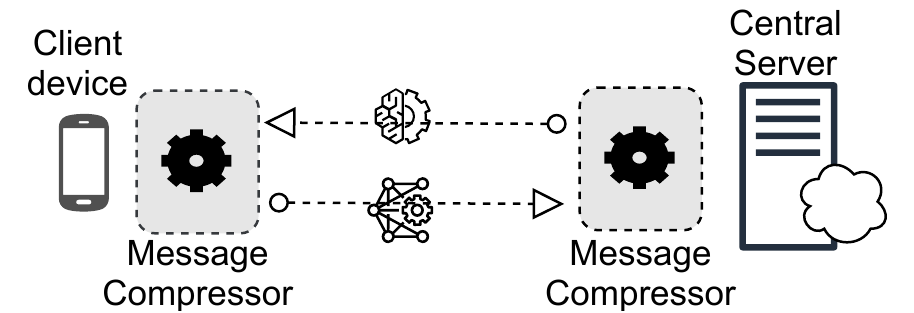}

\caption{Message Compressor.}
\label{Fig:MessageCompressor}
\end{figure}

\textbf{Summary:} A message compressor reduces the message data size before every round of model exchange to increase the communication efficiency. Fig.~\ref{Fig:MessageCompressor} illustrates the operation of the pattern on both ends of the system (client device and central server).

\textbf{Context:} Multiple rounds of model exchanges occurs between a central server and many client devices to complete the model training.

\textbf{Problem:} Communication cost for model communication (e.g., transferring model parameters or gradients) is often a critical bottleneck when the system scales up, especially for bandwidth-limited client devices. 
 
\textbf{Forces:} The problem requires to balance the following forces:

\begin{itemize}[leftmargin=*]
\item \textit{Computation cost.} High computation costs are required by the central server to aggregate all the bulky model parameters collected every round.
\item \textit{Communication cost.} Communication costs are scarce to communicate the model parameters and gradients between resource-limited client devices and the central server. 

\end{itemize}

\textbf{Solution:} The model parameters and the training task script as one message package is compressed before being transferred between the central server and client devices.

\textbf{Consequences:}

Benefits:
\begin{itemize}[leftmargin=*]
\item  \textit{Communication efficiency.} The compression of model parameters reduces the communication cost and network\\ throughput for model exchanges.

\end{itemize}

Drawbacks: 
\begin{itemize}[leftmargin=*]
\item  \textit{Computation cost.} Extra computation is required for message compression and decompression every round. 

\item \textit{Loss of information.} The downsizing of the model parameters might cause the loss of essential information. 
\end{itemize}

\textbf{Related patterns:} \textit{Client Registry, Model Co-Versioning Registry}

\textbf{Known uses:}
\begin{itemize}[leftmargin=*]
\item \textit{Google Sketched Update}~\cite{konecny2017federated}: Google proposes two communication efficient update approaches: structured update and sketched update. Structured update directly learns an update from a restricted space that can be parametrised using a smaller number of variables, whereas sketched update compresses the model before sending it to the central server.

\item \textit{IBM PruneFL}~\cite{jiang2020model} adaptively prunes the distributed parameters of the models, including initial pruning at a selected client and further pruning as part of the federated learning process.

\item \textit{FedCom}~\cite{haddadpour2020federated} compresses messages for uplink communication from the client device to the central server. The central server produces a convex combination of the previous global model and the average of updated local models to retain the essential information of the compressed model parameters.

\end{itemize}

\subsubsection{\textbf{Pattern 5: Model Co-versioning Registry}}\label{ModelCo-versioningRegistry}

\begin{figure}[h]
\centering\includegraphics[width=0.75\linewidth]{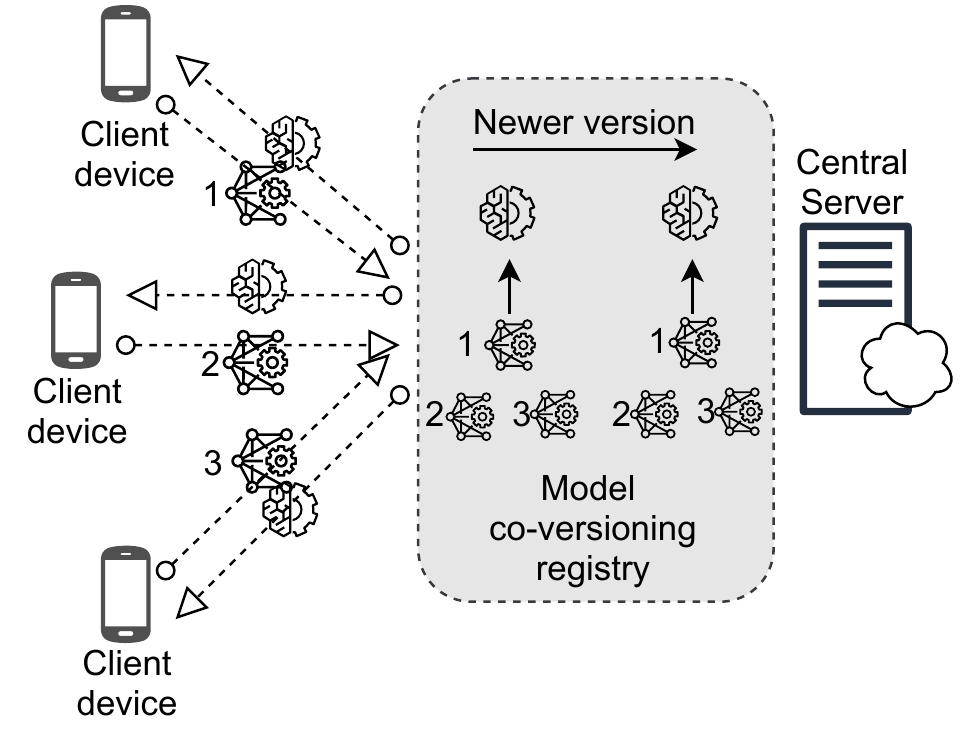}

\caption{Model Co-versioning Registry.}
\label{Fig:ModelCo-versioningRegistry}
\end{figure}

\textbf{Summary:} A model co-versioning registry records all local model versions from each client and aligns them with their corresponding global model. This enables the tracing of model quality and adversarial client devices at any specific point of the training to improve system accountability. Fig.~\ref{Fig:ModelCo-versioningRegistry} shows that the registry collects and maps the local model updates to the associated global model versions.

\textbf{Context:} Multiple new versions of local models are generated from different client devices and one global model aggregated each round. For instance, a federated learning task that runs for 100 rounds on 100 devices will create 10,000 local models and 100 global models in total.    

\textbf{Problem:} With high numbers of local models created each round, it is challenging to keep track of which local models contributed to the global model of a specific round. Furthermore, the system needs to handle the challenges of asynchronous updates, client dropouts, model selection, etc.	

\textbf{Forces:} The problem requires to balance the following forces:

\begin{itemize}[leftmargin=*]
    \item \textit{Updatability.} The system needs to keep track of the local and global models concerning each client device's updates (application's version or device OS/firmware updates) and ensure that the information recorded is up-to-date.

    \item \textit{Immutability.} The records and storage of the models co-versions and client IDs needs to be immutable.

\end{itemize}

\textbf{Solution:} A model co-versioning registry records the local model version of each client device and the global model it corresponds to. This enables seamless synchronous and asynchronous model updates and aggregation. Furthermore, the model co-versioning registry enables the early-stopping of complex model training (stop training when the local model overfits and retrieve the best performing model previously). This can be done by observing the performance of the aggregated global model. Moreover, to provide accountable model provenance and co-versioning, blockchain is considered as one alternative to the central server due to immutability and decentralisation properties.

\textbf{Consequences:}

Benefits:
\begin{itemize}[leftmargin=*]
    \item \textit{Model quality.} The mapping of local models with their corresponding version of the global model allows the study of the effect of each local model quality on the global model. 
    
    \item \textit{Accountability.} System accountability improves as stakeholders can trace the local models that correspond to the current or previous global model versions.
    
    \item \textit{System security.} It enables the detection of adversarial or dishonest clients and tracks the local models that poisons the global model or causes system failure. 
    
\end{itemize}

Drawbacks:

\begin{itemize}[leftmargin=*]
    \item \textit{Storage cost.} Extra storage cost is incurred to store all the local and global models and their mutual relationships. The record also needs to be easily retrievable and it is challenging if the central server host more task.
    
\end{itemize}

\textbf{Related patterns:}~\textit{Client Registry}

\textbf{Known uses:}
\begin{itemize}[leftmargin=*]
 \item \textit{DVC}\footnote{\url{https://dvc.org/}} is an online machine learning version control platform built to make models shareable and reproducible. 

\item \textit{MLflow Model Registry}\footnote{\url{https://docs.databricks.com/applications/mlflow/model-registry.html}} on Databricks is a centralized model store that provides chronological model lineage, model versioning, stage transitions, and descriptions.

\item \textit{Replicate.ai}\footnote{\url{https://replicate.ai/}} is an open-source version control platform for machine learning that automatically tracks code, hyperparameters, training data, weights, metrics, and system dependencies.

\item \textit{Pachyderm}\footnote{\url{https://www.pachyderm.com/}} is an online machine learning pipeline platform that uses containers to execute the different steps of the pipeline and also solves the data versioning provenance issues.

\end{itemize}

\subsubsection{\textbf{Pattern 6: Model Replacement Trigger}}\label{ModelReplacementTrigger}

\begin{figure}[h]
\centering\includegraphics[width=0.7\linewidth]{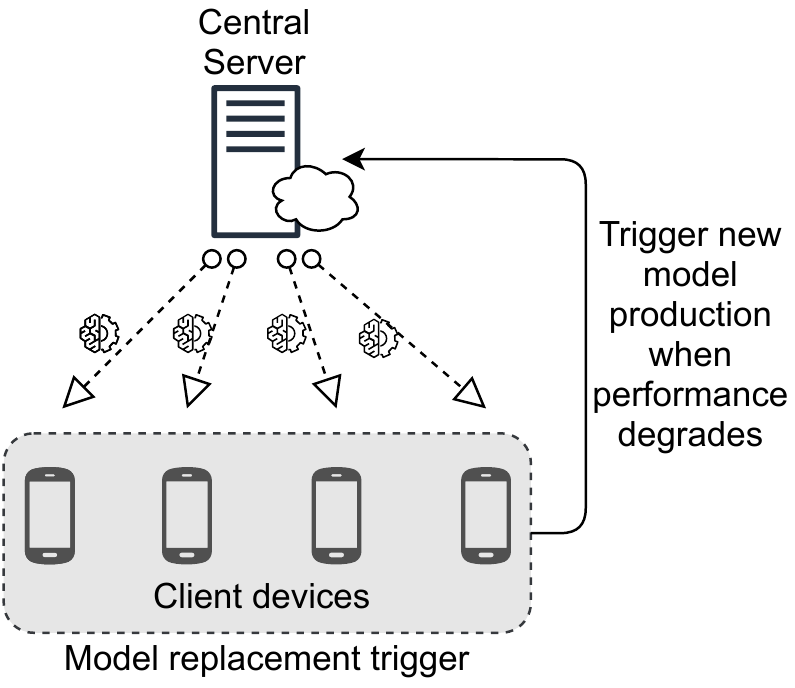}

\caption{Model Replacement Trigger.}
\label{Fig:ModelReplacementTrigger}
\end{figure}

\textbf{Summary:} Fig.~\ref{Fig:ModelReplacementTrigger} illustrates a model replacement trigger that initiates a new model training task when the current global model's performance drops below the threshold value or when a degrade on model prediction accuracy is detected.   	    

\textbf{Context:} The client devices use converged global models for inference or prediction.	  	  	  	     

\textbf{Problem:} As new data is introduced to the system, the global model accuracy might reduce gradually. Eventually, with the degrading performance, the model is no longer be suitable for the application.	

\textbf{Forces:} The problem requires to balance the following forces:

\begin{itemize}[leftmargin=*]
    \item \textit{Model quality.} The global model deployed might experience a performance drop when new data are used for inference and prediction.

    \item \textit{Performance degradation detection.} The system needs to effectively determine the reason for the global model's performance degradation before deciding whether to activate a new global model generation.
\end{itemize}

\textbf{Solution:} A model replacement trigger initiates a new model training task when the acting global model's performance drops below the threshold value. It will compare the performance of the deployed global model on a certain number of client devices to determine if the degradation is a global event. When the global model performance is lower than the preset threshold value for more than a fixed number of consecutive times, given that performance degradation is a global event, a new model training task is triggered.
 
\textbf{Consequences: }

Benefits:
\begin{itemize}[leftmargin=*]
    \item \textit{Updatability.} The consistent updatability of the global model helps to maintain system performance and reduces the non-IID data effect. It is especially effective for clients that generate highly personalised data that causes the effectiveness of the global model to reduce much faster as new data is generated. 
   
    \item \textit{Model quality.} The ongoing model performance monitoring is effective to maintain the high quality of the global model used by the clients.
    
\end{itemize}

Drawbacks:

\begin{itemize}[leftmargin=*]
    \item \textit{Computation cost.} The client devices will need to perform model evaluation periodically that imposes extra computational costs. 
    
    \item \textit{Communication cost.} The sharing of the evaluation results among clients to know if performance degradation is a global event is communication costly.
    
\end{itemize}

\textbf{Related patterns:} \textit{Client Registry, Client Selector, Model Co-versioning Registry}

\textbf{Known uses: }
\begin{itemize}[leftmargin=*]
 \item \textit{Microsoft Azure Machine Learning Designer}\footnote{\url{https://azure.microsoft.com/en-au/services/machine-learning/designer/}} provides a platform for machine learning pipeline creation that enables models to be retrained on new data.

\item \textit{Amazon SageMaker}\footnote{\url{https://aws.amazon.com/sagemaker/}} provides model deployment and monitoring services to maintain the accuracy of the deployed models.

\item \textit{Alibaba Machine Learning Platform}\footnote{\url{https://www.alibabacloud.com/product/machine-learning}} provides end-to-end machine learning services, including data processing, feature engineering, model training, model prediction, and model evaluation.

\end{itemize}

\subsubsection{\textbf{Pattern 7: Deployment Selector}}\label{DeploymentSelector}

\begin{figure}[h]
\centering\includegraphics[width=0.68\linewidth]{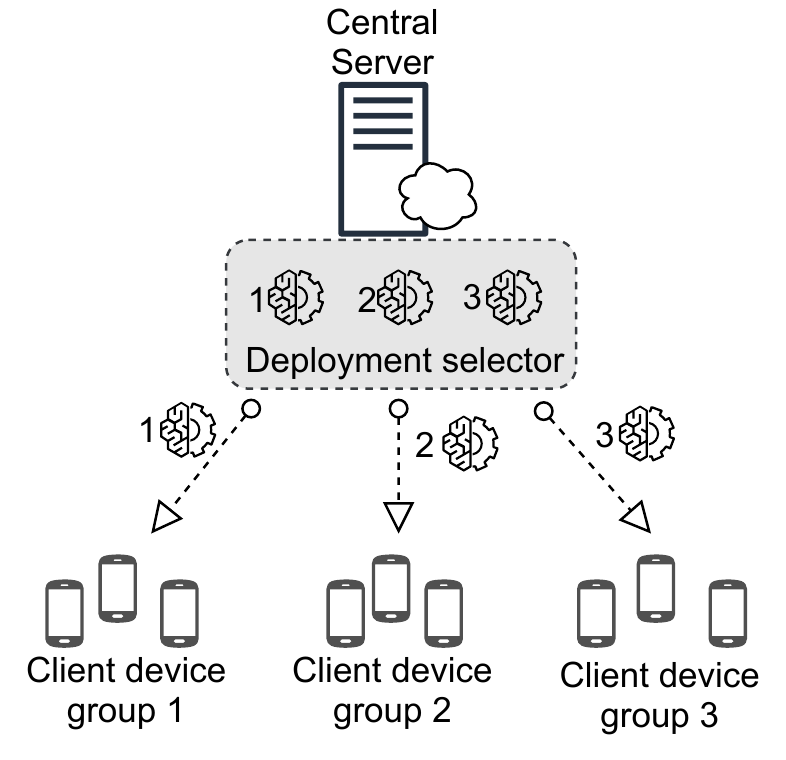}

\caption{Deployment Selector.}
\label{Fig:DeploymentSelector}
\end{figure}

\textbf{Summary:} A deployment selector deploys the converged global model to the selected model users to improve the prediction quality for different applications and tasks. As shown in Fig.~\ref{Fig:DeploymentSelector}, different versions of converged models are deployed to different groups of clients after evaluation.

\textbf{Context:} Client devices train local models using multitask federated learning settings (a model is trained using data from multiple applications to perform similar and related tasks). These models need to be deployed to suitable groups of client devices.		  	  	  	     

\textbf{Problem:} Due to the inherent diversity and non-IID distribution among local data, the globally trained model may not be accurate enough for all clients or tasks. 	

\textbf{Forces:} The problem requires to balance the following forces:

\begin{itemize}[leftmargin=*]
    \item \textit{Identification of clients.} The central server needs to match and deploy the global models to the different groups of client devices.

    \item \textit{Training and storage of different models.} The central server needs to train and store different global models for diverse clients or applications.

\end{itemize}

\textbf{Solution:} A deployment selector examines and selects clients (i.e., model users) to receive the trained global model specified for them based on their data characteristics or applications. The deployment selector deploys the model to the client devices once the global model is completely trained.

\textbf{Consequences:}

Benefits:
\begin{itemize}[leftmargin=*]
    \item \textit{Model performance.} Deploying converged global models to suitable groups of client devices enhances the model performance to the specific groups of clients or applications.

\end{itemize}

Drawbacks:

\begin{itemize}[leftmargin=*]
    \item \textit{Cost.} There are extra costs for training of multiple personalised global models, deployment selection, storage of multiple global models.
    
    \item \textit{Model performance.} The statistical heterogeneity of model trainers produces personalised local models, which is then generalised through~\textit{FedAvg} aggregation. We need to consider the performance trade-off of the generalised global model deployed for different model users and applications. 
    
    \item \textit{Data privacy.} Data privacy challenges exist when the central server collects the client information to identify suitable models for different clients. Moreover, the global model might be deployed to model users that have never joined the model training process.

\end{itemize}

\textbf{Related patterns:} \textit{Client Registry, Client Selector}

\textbf{Known uses: }
\begin{itemize}[leftmargin=*]
 \item \textit{Azure Machine Learning}\footnote{\url{https://docs.microsoft.com/en-us/azure/machine-learning/concept-model-management-and-deployment}} supports mass deployment with a step of compute target selection.
 \item \textit{Amazon SageMaker}\footnote{\url{https://docs.aws.amazon.com/sagemaker/latest/dg/multi-model-endpoints.html}} can host multiple models with multi-model endpoints.
 \item \textit{Google Cloud}\footnote{\url{https://cloud.google.com/ai-platform/prediction/docs/deploying-models}} uses model resources to manage different versions of models.

\end{itemize}

\subsection{\textbf{Model Training Patterns}}\label{ModelTrainingPatterns}

Patterns about the model training and data preprocessing are group together as model training patterns, including \textit{multi-task model trainer} that tackles non-IID data characteristics, \textit{heterogeneous data handler} that deals with data heterogeneity in training datasets, and \textit{incentive registry} that increases the client's motivatability through rewards.

\subsubsection{\textbf{Pattern 8: Multi-Task Model Trainer}}\label{Multi-taskModelTrainer}

\begin{figure}[h]
\centering\includegraphics[width=0.63\linewidth]{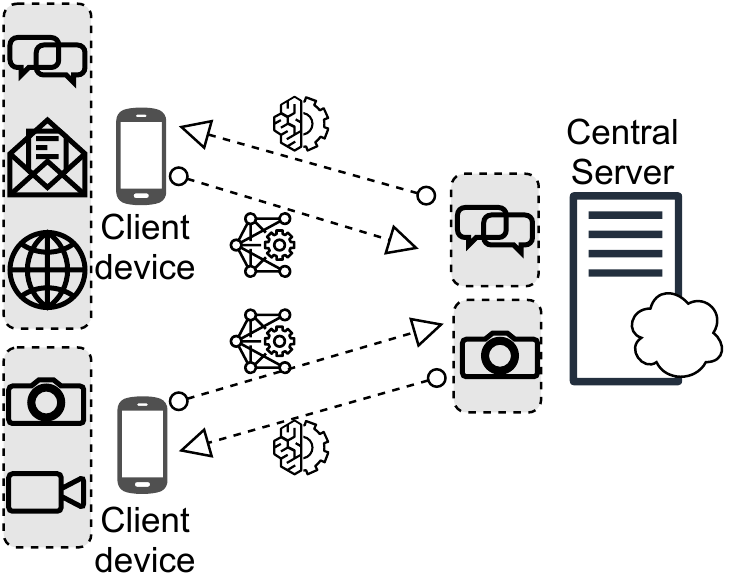}

\caption{Multi-Task Model Trainer.}
\label{Fig:Multi-taskModelTrainer}
\end{figure}

\textbf{Summary:} In federated learning, a multi-task model trainer trains separated but related models on local client devices to improve learning efficiency and model performance. As shown in Fig.~\ref{Fig:Multi-taskModelTrainer}, there are two groups of applications: (i) text-related applications (e.g., messaging, email, etc.), and (ii) image-related applications (camera, video, etc.). The related models are trained on client devices using data of related tasks.

\textbf{Context:} Local data has statistical heterogeneity property where the data distribution among different clients is skewed and a global model cannot capture the data pattern of each client.

\textbf{Problem:} Federated learning models trained with non-IID data suffer from low accuracy and are less generalised to the entire dataset. Furthermore, the local data that is highly personalised to the device users' usage pattern creates local models that diverge in different directions. Hence, the global model may have relatively low averaged accuracy.	  

\textbf{Forces:} The problem requires to balance the following forces:

\begin{itemize}[leftmargin=*]
    \item \textit{Computation cost.} The complex model that solves the non-IID issue consumes more computation and energy resources every round compared to general federated model training. It also takes longer to compute all the training results from the different tasks before submitting them to the central server.

    \item \textit{Data privacy.} To address the non-IID issue, more information from the local data needs to be explored to understand the data distribution pattern. This ultimately exposes client devices to local data privacy threats.	 

\end{itemize}

\textbf{Solution:} The multi-task model trainer performs similar-but-related machine learning tasks on client devices. This enables the local model to learn from more local data that fit naturally to the related local models for different tasks. For instance, a multi-task model for the next-word prediction task is trained using the on-device text messages, web browser search strings, and emails with similar mobile keyboard usage patterns. \textit{MOCHA}~\cite{smith2018federated} is a state-of-the-art federated multi-task learning algorithm that realises distributed multi-task learning on federated settings. 

\textbf{Consequences:}

Benefits:
\begin{itemize}[leftmargin=*]
    \item \textit{Model quality.} Multi-task learning improves the model performance by considering local data and loss in optimization and obtaining a local weight matrix through this process. The local model fits for non-IID data in each node better than a global model. 
    
\end{itemize}

Drawbacks:

\begin{itemize}[leftmargin=*]
    
    \item \textit{Model quality.} Multi-task training often works only with convex loss functions and performs weak on non-convex loss functions. 
    
    \item \textit{Model portability.} As each client has a different model, the model's portability is a problem that makes it hard to apply multi-task training on cross-device FL.
\end{itemize}

\textbf{Related patterns:} \textit{Client Registry, Model Co-versioning Registry, Client Cluster, Deployment Selector}

\textbf{Known uses: }
\begin{itemize}[leftmargin=*]
 
 \item \textit{MultiModel}\footnote{\url{https://ai.googleblog.com/2017/06/multimodel-multi-task-machine-learning.html}} is a neural network architecture by Google that simultaneously solves several problems spanning multiple domains, including image recognition, translation, and speech recognition.
 
 \item \textit{MT-DNN}\footnote{\url{https://github.com/microsoft/MT-DNN}} is an open-source natural language understanding toolkit by Microsoft to train customized deep learning models. 
 
 \item \textit{Yahoo Multi-Task Learning for Web Ranking} is a multi-task learning framework developed by Yahoo! Labs to rank in web search.
 
 \item \textit{VIRTUAL}~\cite{corinzia2019variational} is an algorithm for federated multi-task learning with non-convex models. The server and devices are treated as a star-shaped bayesian network, and model learning is performed on the network using approximated variational inference.

\end{itemize}

\subsubsection{\textbf{Pattern 9: Heterogeneous Data Handler}}\label{HeterogeneousDataHandler}

\begin{figure}[h]
\centering\includegraphics[width=0.75\linewidth]{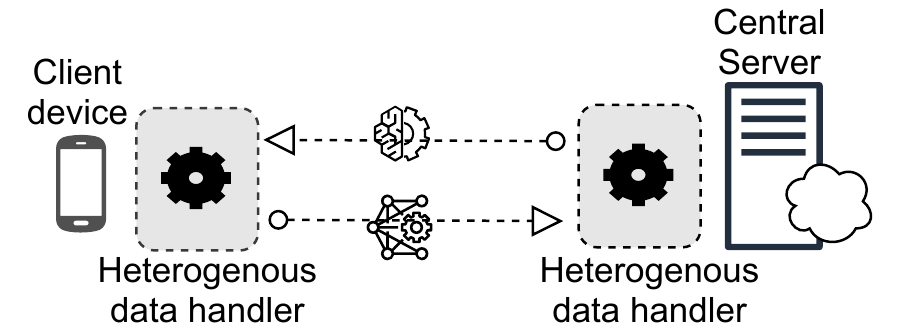}

\caption{Heterogeneous Data Handler.}
\label{Fig:HeterogeneousDataHandler}
\end{figure}

\textbf{Summary:} Heterogeneous data handler solves the non-IID and skewed data distribution issues through data volume and data class addition (e.g., data augmentation or generative adversarial network) while maintaining the local data privacy. The pattern is illustrated in Fig.~\ref{Fig:HeterogeneousDataHandler}, where the heterogeneous data handler operates at both ends of the system. 	   

\textbf{Context:} Client devices possess heterogeneous data characteristics due to the highly personalized data generation pattern. Furthermore, the raw local data cannot be shared so the data balancing task becomes extremely challenging.	  	  	  	  	     

\textbf{Problem:} The imbalanced and skewed data distribution of client devices produces local models that are not generalised to the entire client network. The aggregation of these local models reduces global model accuracy. 	

\textbf{Forces:} The problem requires the following forces to be balanced:


\begin{itemize}[leftmargin=*]
    \item \textit{Data efficiency.} It is challenging to articulate the suitable data volume and classes to be augmented to solve data heterogeneity on local client devices. 
    
    \item \textit{Data accessibility.} The heterogeneous data issue that exists within the client device can be solved by collecting all the data under a centralized location. However, this violates the data privacy of client devices.

\end{itemize}

\textbf{Solution:} A heterogeneous data handler balances the data distribution and solves the data heterogeneity issue in the client devices through data augmentation and federated distillation. Data augmentation solves data heterogeneity by generating augmented data locally until the data volume is the same across all client devices. Furthermore, the classes in the datasets are also populated equally across all client devices. Federated distillation enables the client devices to obtain knowledge from other devices periodically without directly accessing the data of other client devices. Other methods includes taking the quantified data heterogeneity weightage (e.g, Pearson's correlation, centroid averaging-distance, etc.) into account for model aggregation.

\textbf{Consequences:}

Benefits:
\begin{itemize}[leftmargin=*]
    \item \textit{Model quality.} By solving the non-IID issue of local datasets, the performance and generality of the global model are improved.

\end{itemize}

Drawbacks:

\begin{itemize}[leftmargin=*]
    
    \item \textit{Computation cost.} It is computationally costly to deal with data heterogeneity together with the local model training.

\end{itemize}

\textbf{Related patterns:} \textit{Client Registry, Client Selector, Client Cluster}

\textbf{Known uses:}
\begin{itemize}[leftmargin=*]
  \item \textit{Astreae}~\cite{8988732} is a self-balancing federated learning framework that alleviates the imbalances by performing global data distribution-based data augmentation. 
  
   \item Federated Augmentation (\textit{FAug})~\cite{jeong2018communicationefficient} is a data augmentation scheme that utilises a generative adversarial network (GAN) which is collectively trained under the trade-off between privacy leakage and communication overhead. 
   
   \item Federated Distillation (\textit{FD})~\cite{8904164} is a method that adopted knowledge distillation approaches to tackle the non-IID issue by obtaining the knowledge from other devices during the distributed training process, without accessing the raw data. 
\end{itemize}

\subsubsection{\textbf{Pattern 10: Incentive Registry}}\label{IncentiveRegistry}

\begin{figure}[h]
\centering\includegraphics[width=0.6\linewidth]{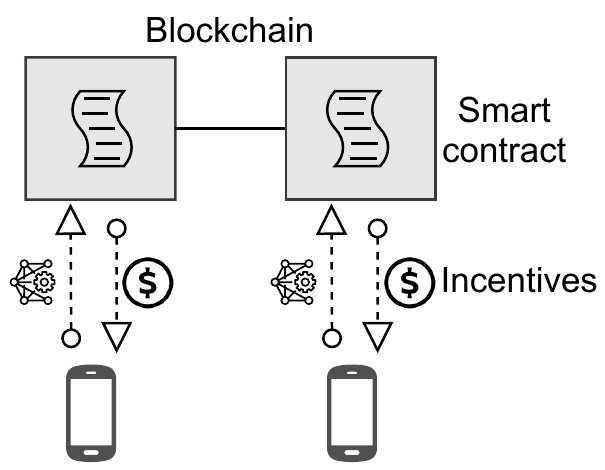}

\caption{Incentive Registry.}
\label{Fig:IncentiveRegistry}
\end{figure}

\textbf{Summary:} An incentive registry maintains the list of participating clients and their rewards that correspond to clients' contributions (e.g., data volume, model performance, computation resources, etc.) to motivate clients' participation. Fig.~\ref{Fig:IncentiveRegistry} illustrates a blockchain \& smart contract-based incentive mechanism. 

\textbf{Context:} The model training participation rate of client devices is low while the high participation rate is crucial for the global model performance. 	     

\textbf{Problem:} Although the system relies greatly on the participation of clients to produce high-quality global models, clients are not mandatory to join the model training and contribute their data/resources. 

\textbf{Forces:} The problem requires to balance the following forces:

\begin{itemize}[leftmargin=*]
    
    \item \textit{Incentive scheme.} It is challenging to formulate the form of rewards to attract different clients with different participation motives. Furthermore, the incentive scheme needs to be agreed upon by both the learning coordinator and the clients, e.g., performance-based, data-contribution-based, resource-contribution-based, and provision-frequency-\\based.
    
    \item \textit{Data privacy.} To identify the contribution of each client device, the local data and client information is required to be studied and analysed by the central server. This exposes the client devices' local data to privacy threats.

\end{itemize}

\textbf{Solution:} An incentive registry records all client's contributions and incentives based on the rate agreed. There are various ways to formulate the incentive scheme, e.g., deep reinforcement learning, blockchain/smart contracts, and the Stackelberg game model.   	  
 
\textbf{Consequences:}

Benefits:
\begin{itemize}[leftmargin=*]
    \item \textit{Client motivatability.} The incentive mechanism is effective in attracting clients to contribute data and resources to the training process.

\end{itemize}

Drawbacks:

\begin{itemize}[leftmargin=*]
    \item \textit{System security.} There might be dishonest clients that submit fraudulent results to earn rewards illegally and harm the training process.

\end{itemize}

\textbf{Related patterns:} \textit{Client Registry, Client Selector}

\textbf{Known uses:}
\begin{itemize}[leftmargin=*]
 \item \textit{FLChain}~\cite{8905038} is a federated learning blockchain providing an incentive scheme for collaborative training and a market place for model trading. 
 
\item \textit{DeepChain}~\cite{8894364} is a blockchain-based collaborative training framework with an incentive mechanism that encourages parties to participate in the deep learning model training and share the obtained local gradients. 

\item \textit{FedCoin}~\cite{Liu2020} is a blockchain-based peer-to-peer payment system for federated learning to enable Shapley Value (SV) based reward distribution. 
 
\end{itemize}

\subsection{\textbf{Model Aggregation Patterns}}\label{ModelAggregationPatterns}

Model aggregation patterns are design solutions of model aggregation used for different purposes. \textit{Asynchronous aggregator} aims to reduce aggregation latency and increase system efficiency, whereas \textit{decentralised aggregator} targets to increase system reliability and accountability. \textit{Hierarchical aggregator} is adopted to improve model quality and optimises resources. \textit{Secure aggregator} is designed to protect the models' security. 

\subsubsection{\textbf{Pattern 11: Asynchronous Aggregator}}\label{AsynchronousAggregator}

\begin{figure}[h]
\centering
\includegraphics[width=0.7\linewidth]{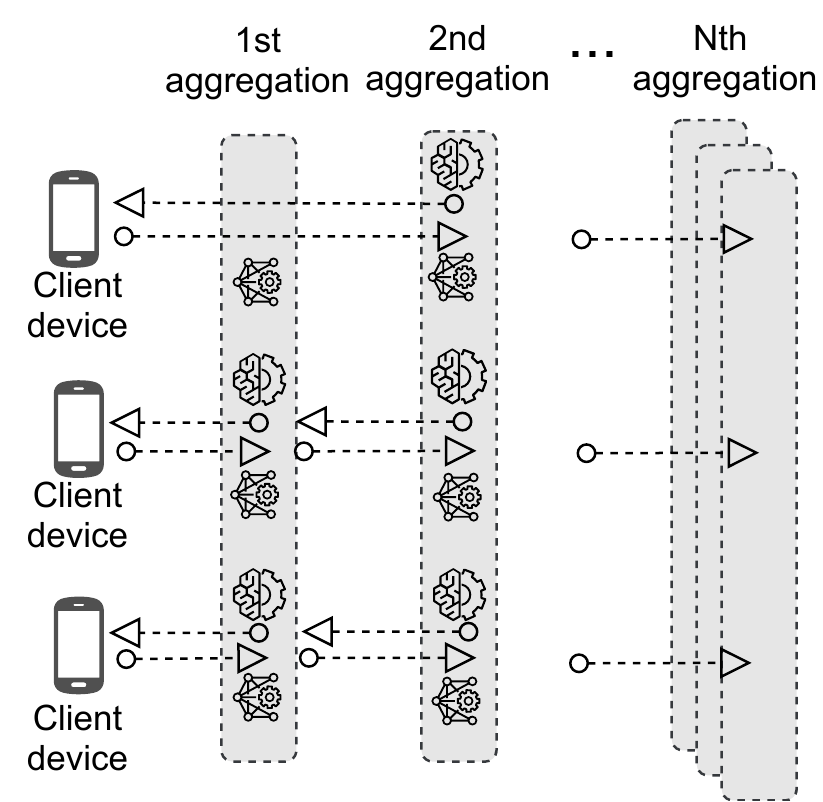}

\caption{Asynchronous Aggregator.}
\label{Fig:AsynchronousAggregator}
\end{figure}

\textbf{Summary:} To increase the global model aggregation speed every round, the central server can perform model aggregation asynchronously whenever a model update arrives without waiting for all the model updates every round. In Fig.~\ref{Fig:AsynchronousAggregator}, the asynchronous aggregator pattern is demonstrated as the first client device asynchronously uploads its local model during the second aggregation round while skipping the first aggregation round.

\textbf{Context:} In conventional federated learning, the central server receives all local model updates from the distributed client devices synchronously and performs model aggregation every round. The central server needs to wait for every model to arrive before performing the model aggregation for that round. Hence, the aggregation time depends on the last model update that reaches the central server.  

\textbf{Problem:} Due to the difference in computation resources, the model training lead time is different per device. Furthermore, the difference in bandwidth availability, communication efficiency affects the model's transfer rate. Therefore, the delay in model training and transfer increases the latency in global model aggregation. 

\textbf{Forces:} The problem requires the following forces to be balanced:

\begin{itemize}[leftmargin=*]
    \item \textit{Model quality.} There will be possible bias in the global model if not all local model updates are aggregated in every iteration as important information or features might be left out.
    
    \item \textit{Aggregation latency.} The aggregation of local models can only be performed when all the model updates are collected. Therefore, the latency of the aggregation process is affected by the slowest model update that arrives at the central server.

\end{itemize}

\textbf{Solution:} The global model aggregation is conducted whenever a model update is received, without being delayed by other clients. Then the server starts the next iteration and distributes the new central model to the clients that are ready for training. The delayed model updates that are not included in the current aggregation round will be added in the next round with some reduction in the weightage, proportioned to their respective delayed time.

\textbf{Consequences:}

Benefits:
\begin{itemize}[leftmargin=*]
    \item \textit{Low aggregation latency.} Faster aggregation time per round is achievable as there is no need to wait for the model updates from other clients for the aggregation round. The bandwidth usage per iteration is reduced as fewer local model updates are transferred and receive simultaneously every round. 
    
\end{itemize}

Drawbacks:

\begin{itemize}[leftmargin=*]
    \item \textit{Communication cost.} The number of iteration to collect all local mode updates increases for the asynchronous approach. More iterations are required for the entire training process to train the model till convergence compares to synchronous global model aggregation.
     \item \textit{Model bias.} The global model of each round does not contain all the features and information of every local model update. Hence the global model might have a certain level of bias in prediction.
    
\end{itemize}

\textbf{Related patterns:} \textit{Client Registry, Client Selector, Model Co-versioning Registry, Client Update Scheduler}

\textbf{Known uses:}
\begin{itemize}[leftmargin=*]
 \item Asynchronous Online Federated Learning (\textit{ASO-fed})~\cite{chen2020asynchronous} is a framework for federated learning that adopted asynchronous aggregation. The central server update the global model whenever it receives a local update from one client device (or several client devices if the local updates are received simultaneously). On the client device side, online-learning is performed as data continue to arrive during the global iterations.
 
 \item Asynchronous federated SGD-Vertical Partitioned (\textit{AFSGD-VP})~\cite{gu2020privacypreserving} algorithm uses a tree-structured communication scheme to perform asynchronous aggregation. The algorithm does not need to align the iteration number of the model aggregation from different client devices to compute the global model.
 
 \item Asynchronous Federated Optimization (\textit{FedAsync})~\cite{xie2020asynchronous} is an approach that leverages asynchronous updating technique and avoids server-side timeouts and abandoned rounds while requires no synchronous model broadcast to all the selected client devices.
\end{itemize}

\subsubsection{\textbf{Pattern 12: Decentralised Aggregator}}\label{DecentralisedAggregator}

\begin{figure}[h]
\centering\includegraphics[width=0.65\linewidth]{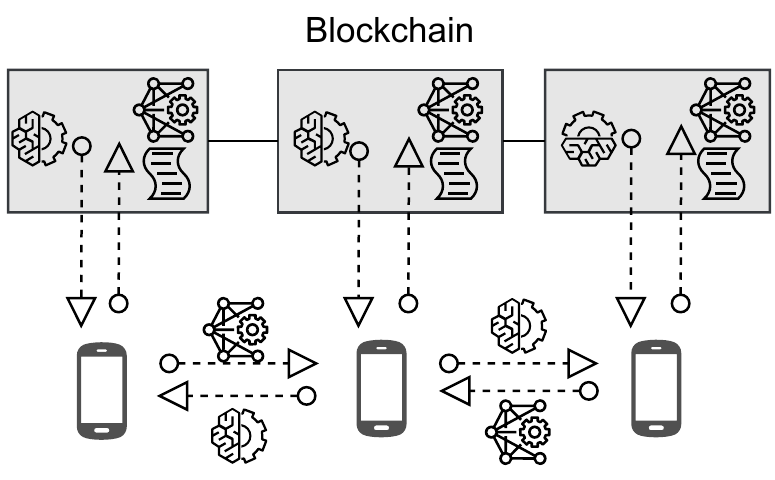}

\caption{Decentralised Aggregator.}
\label{Fig:DecentralisedAggregator}
\end{figure}

\textbf{Summary:} A decentralised aggregator improves system reliability and accountability by removing the central server that is a possible single-point-of-failure. Fig.~\ref{Fig:DecentralisedAggregator} illustrates the decentralised federated learning system built using blockchain and smart contract, while the model updates are performed through the exchange between neighbour devices.

\textbf{Context:} The model training and aggregation are coordinated by a central server and both the central server and the owner may not be trusted by all the client devices that join the training process.	  	

\textbf{Problem:} In~\textit{FedAvg}, all the chosen devices have to submit the model parameters to one central server every round. This is extremely burdensome to the central server and network congestion may occur. Furthermore, centralised federated learning possesses a single-point-of-failure. Data privacy threats may also occur if the central server is compromised by any unauthorised entity. The mutual trust between the client devices and the central server may not be specifically established.

\textbf{Forces: }The problem requires to balance the following forces:

\begin{itemize}[leftmargin=*]
    \item \textit{Decentralised model management.} The federated learning systems face challenges to collect, store, examine, and aggregate the local models due to the removal of the central server.
    
    \item \textit{System ownership.} Currently, the central server is own by the learning coordinator that creates the federated learning jobs. The removal of the central server requires the re-definition of system ownership. It includes the authority and accessibility of learning coordinator in the federated learning systems.

\end{itemize}

\textbf{Solution:} A decentralised aggregator replaces the central server's role in a federated learning system. The aggregation and update of the models can be performed through peer-to-peer exchanges between client devices. First, a random client from the system can be an aggregator by requesting the model updates from the other clients that are close to it. Simultaneously, the client devices conduct local model training in parallel and send the trained local models to the aggregator. The aggregator then produces a new global model and sends it to the client network. Blockchain is the alternative to the central server for model storage that prevents single-point-of-failure. The ownership of the blockchain belongs to the learning coordinator that creates the new training tasks and maintains the blockchain. Furthermore, the record of models on a blockchain is immutable that increases the reliability of the system. It also increases the trust of the system as the record is transparent and accessible by all the client devices.

\textbf{Consequences: }

Benefits:
\begin{itemize}[leftmargin=*]
    \item \textit{System reliability.} The removal of single-point-of-failure increases the system reliability by reducing the security risk of the central server from any adversarial attack or the failure of the entire training process due to the malfunction of the central server.
    
    \item \textit{System accountability.} The adoption of blockchain promotes accountability as the records on a blockchain is immutable and transparent to all the stakeholders.
    
\end{itemize}

Drawbacks:

\begin{itemize}[leftmargin=*]
    \item \textit{Latency.} Client device as a replacement of the central server for model aggregation is not ideal for direct communication with multiple devices (star-topology). This may cause latency in the model aggregation process due to blockchain consensus protocols. 
    
    \item \textit{Computation cost.} Client devices have limited computation power and resource to perform model training and aggregation parallel. Even if the training process and the aggregation is performed sequentially, the energy consumption to perform multiple rounds of aggregation is very high. 
    
    \item \textit{Storage cost.} High storage cost is required to store all the local and global models on storage-limited client devices or blockchain.
    
    \item \textit{Data privacy.} Client devices can access the record of all the models under decentralised aggregation and blockchain settings. This might expose the privacy-sensitive information of the client devices to other parties.	  
    
\end{itemize}

\textbf{Related patterns:} \textit{Model Co-versioning Registry, Incentive Registry}

\textbf{Known uses:}
\begin{itemize}[leftmargin=*]
 \item \textit{BrainTorrent}~\cite{roy2019braintorrent} is a server-less, peer-to-peer approach to perform federated learning where clients communicate directly among themselves, specifically for federated learning in medical centers.
 
 \item \textit{FedPGA}~\cite{9205506} is a decentralised aggregation algorithm developed from~\textit{FedAvg}. The devices in~\textit{FedPGA} exchange partial gradients rather than full model weights. The partial gradient exchange pulls and merges the different slice of the updates from different devices and rebuild a mixed update for aggregation.
 
 \item A fully decentralised framework~\cite{lalitha2018fully} is an algorithm in which users update their beliefs by aggregate information from their one-hop neighbors to learn a model that best fits the observations over the entire network.
 
 \item A Segmented gossip approach~\cite{hu2019decentralized} splits a model into segmentation that contains the same number of non-overlapping model parameters. Then, the gossip protocol is adopted where each client stochastically selects a few other clients to exchange the model segmentation for each training iteration without the orchestration of a central server.

\end{itemize}

\subsubsection{\textbf{Pattern 13: Hierarchical Aggregator}}\label{HierarchicalAggregator}

\begin{figure}[h]
\centering\includegraphics[width=0.8\linewidth]{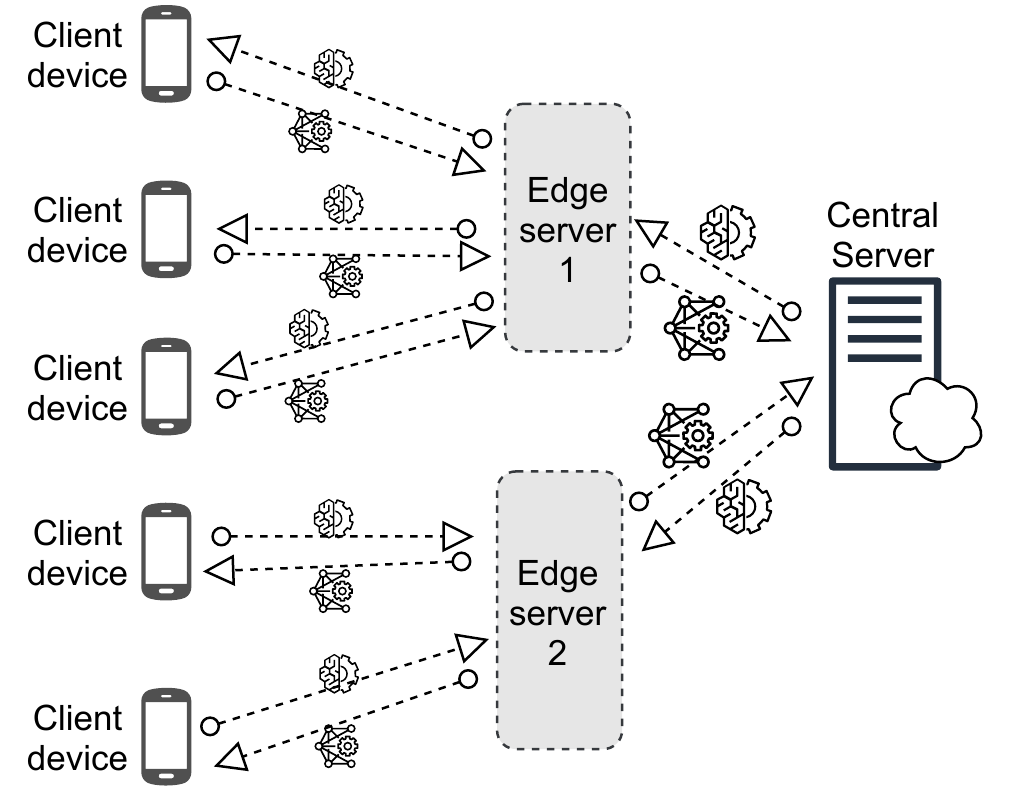}

\caption{Hierarchical Aggregator.}
\label{Fig:HierarchicalAggregator}
\end{figure}

\textbf{Summary:} To reduce non-IID effects on the global model and increase system efficiency, a hierarchical aggregator adds an intermediate layer (e.g., edge server) to perform partial aggregations using the local model parameters from closely-related client devices before the global aggregation. In Fig.~\ref{Fig:HierarchicalAggregator}, edge servers are added as an intermediate layer between the central server and client devices to serve the client devices that are closer to them.  

\textbf{Context:} The communication between the central server and the client devices is slowed down or frequently disrupted due to being physically distant from each other and are wirelessly connected.   	  	     

\textbf{Problem:} The central server can access and store more data but requires high communication overhead and suffers from latency due to being physically distant from the client devices. Moreover, client devices possess non-IID characteristics that affect global model performance.  	

\textbf{Forces:} The problem requires the following forces to be balanced:

\begin{itemize}[leftmargin=*]
    \item \textit{System efficiency.} The system efficiency of the server-client setting to perform federated learning is low, as the central server is burdensome to accommodate the communication and the model aggregations of the widely-distributed client devices. 
    
    \item \textit{Data heterogeneity.} In the server-client setting of a federated learning system, the data heterogeneity characteristics of client devices become influential and dominant to the global model production, as the central server deals with all the client devices that generate non-IID data.
\end{itemize}

\textbf{Solution:} A hierarchical aggregator adds edge servers between the central server and client devices. The combination of server-edge-client architecture can improve both computation and communication efficiency of the federated model training process. Edge servers collect local models from the nearest client devices, a subset of all the client devices. After every k1 round of local training on each client, each edge server aggregates its clients’ models. After every k2 edge model aggregations, the cloud server aggregates all the edge servers’ models, which means the communication with the central server happens every k1k2 local updates~\cite{9148862}.

\textbf{Consequences: }

Benefits:
\begin{itemize}[leftmargin=*]
    \item \textit{Communication efficiency.} The hierarchical aggregators speed up the global model aggregation and improve communication efficiency.
    \item \textit{Scalability.} Adding an edge layer helps to scale the system by improving the system's ability to handling more client devices.
    \item \textit{Data heterogeneity and non-IID reduction.} The partial aggregation in a hierarchical manner aggregates local models that have similar data heterogeneity and non-IIDness before the global aggregation on the central server. This greatly reduces the effect of data heterogeneity and non-IIDness on global models. 
    
    \item \textit{Computation and storage efficiency.} The edge devices are rich with computation and storage resources to perform partial model aggregation. Furthermore, edge devices are nearer to the client devices which increase the model aggregation and computation efficiency.

\end{itemize}

Drawbacks:

\begin{itemize}[leftmargin=*]
\item \textit{System reliability.} The failure of edge devices may cause the disconnection of all the client devices under those edge servers and affect the model training process, model performance, and system reliability.
    
\item \textit{System security.} Edge servers could become security breach points as they have lower security setups than the central server and the client devices. Hence, they are more prone to network security threats or becoming possible points-of-failure of the system.

\end{itemize}

\textbf{Related patterns:} \textit{Client Registry, Client Cluster, Model Co-versioning Registry}

\textbf{Known uses: }
\begin{itemize}[leftmargin=*]
 \item \textit{HierFAVG} 
 is an algorithm that allows multiple edge servers to perform partial model aggregation incrementally from the collected updates from the client devices.
 
 \item Hierarchical Federated Learning (\textit{HFL}) enables hierarchical model aggregation in large scale networks of client devices where communication latency is prohibitively large due to limited bandwidth. The \textit{HFL} seeks a consensus on the model and uses edge servers to aggregate model updates from client devices that are geographically near.
 
 \item Federated Learning + Hierarchical Clustering (\textit{FL+HC}) is the addition of a hierarchical clustering algorithm to the federated learning system. The cluster is formed according to the data distributions similarity based on the following distance metrics: (1) Manhattan, (2) Euclidean, (3) Cosine distance metrics.
 
 \item \textit{Astraea} is a federated learning framework that tackles non-IID characteristics of federated clients. The framework introduces a mediator to the central server and the client devices to balance the skewed client data distributions. The mediator performs the z-score-based data augmentation and downsampling to relieve the global imbalanced of training data.

\end{itemize}

\subsubsection{\textbf{Pattern 14: Secure Aggregator}}\label{SecureAggregator}

\begin{figure}[h]
\centering\includegraphics[width=0.68\linewidth]{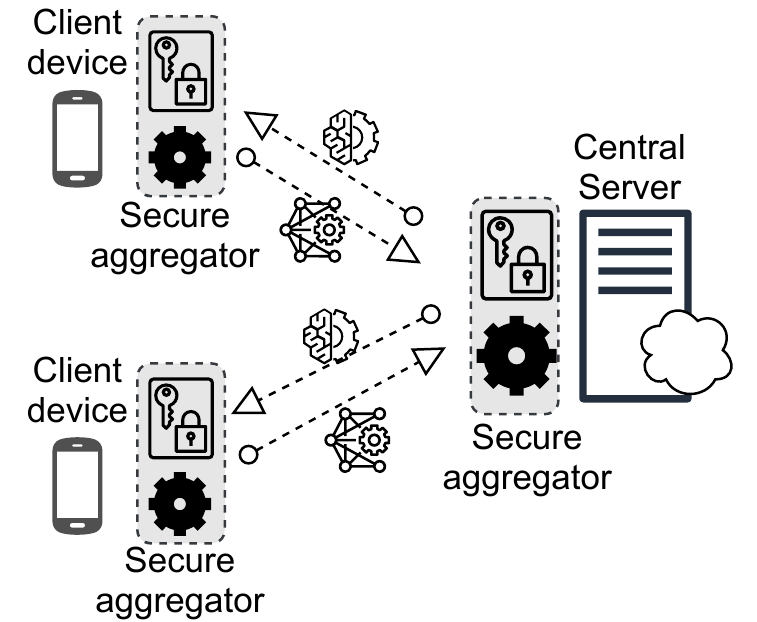}

\caption{Secure Aggregator.}
\label{Fig:SecureAggregator}
\end{figure}

\textbf{Summary:} A security aggregator manages the model exchange and aggregation security protocols to protect model security. Fig.~\ref{Fig:SecureAggregator} illustrates the security aggregator on each components with the security protocols embedded in them.

\textbf{Context:} The central server sends global models to any existing or unknown device every round with no data privacy and security protocols that protect the communication from unauthorised access. Furthermore, model parameters contain pieces of private user information that can be inferred by the data-hungry machine learning algorithms.

\textbf{Problem:} There are no security measures to tackle the honest-but-curious and active adversary security threats which exist in federated learning systems.

\textbf{Forces:} The problem requires to balance the following forces:

\begin{itemize}[leftmargin=*]
    \item \textit{Client device security.} Client device security issues exist when dishonest and malicious client devices join the training process and poison the overall model performance by disrupting the training process or providing false updates to the central server.  
    
    \item \textit{Data security.} Data security of the client devices is challenged when the gradients or model parameters are inferred by unauthorised parties through the data-hungry machine learning algorithms.
 
\end{itemize}

\textbf{Solution:} A security aggregator handles the secure multiparty computation (SMC) protocols for model exchanges and aggregations. The protocols provide security proof to guarantee that each party knows only its input and output. For instance, homomorphic encryption is a method to encrypt the models and only allow authorised client devices and the central server to decrypt and access the models. Pairwise masking and differential privacy (DP) methods are applied to reduce the interpretability of the model by unauthorised party~\cite{10.1145/3298981}. The technique involves adding noise to the parameters or gradient or uses a generalised method.

\textbf{Consequences: }

Benefits:
\begin{itemize}[leftmargin=*]
    \item \textit{Data security.} The secure aggregator protects the model from being access by adversarial and unauthorised parties through homomorphic encryptions and prevents information leakage due to the data-hungry property of machine learning models. 
    
\end{itemize}

Drawbacks:
\begin{itemize}[leftmargin=*]
    \item \textit{System efficiency.} The extra security processes affect the system efficiency if excessive security steps are required every round for every device. It also lowers the training and aggregation speed due to encryption and decryption time. 
    \item \textit{Model performance-privacy trade-off.} The model performance is affected if the model privacy methods aggressively interfere with the model's interpretability due to being excessively obscure. 
    \item \textit{Compromised key.} For encryption and decryption functions, the possible compromise of the security keys increases the privacy threat. 
    	 
\end{itemize}

\textbf{Related patterns:} \textit{Client Registry, Model Co-versioning Registry}

\textbf{Known uses: }
\begin{itemize}[leftmargin=*]
    \item \textit{SecAgg}\cite{10.1145/3133956.3133982} a practical protocol by Google for secure aggregation in the federated learning settings.
    
    \item \textit{HybridAlpha}~\cite{10.1145/3338501.3357371} is a framework that manages the client devices that join the federated learning process. The security operation includes functional encryption, DP, and SMC.

    \item \textit{TensorFlow Privacy Library}\footnote{\url{https://github.com/tensorflow/privacy/}} provides an implementation of DP-SGD machine learning.

    \item \textit{ZamaAI}\footnote{\url{https://zama.ai/}} is an AI service platform that provides encryption technology that uses a homomorphic compiler to convert the model into an end-to-end encrypted parameters.
    
    \item Simple Encrypted Arithmetic Library (\textit{SEAL}\footnote{\url{https://www.microsoft.com/en-us/research/project/microsoft-seal/}}) is a homomorphic encryption API introduced by Microsoft AI to allow computations to be performed directly on encrypted data. 

\end{itemize}

\section{Discussion}
\label{S:discussion}

Various patterns are proposed to improve the architectural design challenges of a federated learning system. The main challenges include communication \& computation efficiency, data privacy, model performance, system security, and reliability. First, \textit{client registry}, \textit{client selector}, and \textit{client cluster} are proposed for client device management in the job creation stage. These patterns manage client devices to improve model performance, system, and training efficiency.

During the model training stage, the performance trade-off often occurs due to the non-IID nature of the local data. The patterns proposed to address this issue are \textit{heterogeneous data handler} and \textit{incentive registry}. Furthermore, the non-IID data that enhances the local personalisation of the model also hurts the generalisation of the global model produced. The patterns proposed to address this issue are \textit{client cluster}, \textit{hierarchical aggregator}, \textit{multi-task model trainer}, and \textit{deployment selector}. Specifically, \textit{multi-task model trainer} adopts multi-task or transfer learning techniques to learn different models or personalise a global model on local data to optimise the model performance for clients with different local data characteristics, whereas the \textit{deployment selector} effectively selects the user clients to receive the personalised models that fit their local data.

For the model exchange and aggregation stages, communication and computation efficiency become a system bottleneck. To effectively tackle these issues, \textit{client selector}, \textit{client cluster}, \textit{deployment selector}, \textit{asynchronous aggregator}, and \textit{hierarchical aggregator} are embedded in the system to optimise resource consumption. However, these patterns require extra client information (i.e., resource or performance) to perform the selection or scheduling of updates. Intuitively, the collection and analysis of the client information on the central server may lead to another form of data privacy violation. Furthermore, extra computation and communication resources are consumed to collect and analyse the client information, in addition to the model training task and the fundamental tasks of the client devices. Hence, \textit{message compressor} and \textit{hierarchical aggregator} are proposed to tackle these issues. Moreover, \textit{incentive registry} is proposed to encourage more client devices to join the training to improve the model performance. 

The system security issue is present due to the distributed ownership of federated learning system components. The client nodes are mostly owned by different parties which are not governed by the system owner. Therefore, unauthorised clients may join the system and obtain model parameters from the system. Furthermore, adversarial clients may harm the model or system performance by uploading dishonest updates. \textit{Secure aggregator}, \textit{model co-versioning registry}, and \textit{client registry} aim to solve these challenges. Lastly, the trustworthiness between the client devices and the central server is also a challenge to gain the participation of clients. The central server may also be a single-point-of-failure that may affect the reliability of the system. Hence, \textit{decentralised aggregator} is proposed to solve the issue. 

\section{Related Work}
\label{S:related}
In many real-world scenarios, machine learning applications are usually embedded as a software component to a larger software system at enterprise level. Hence, to promote enterprise level adoption of machine learning-based applications, many researchers view machine learning models as a component of a software system so that the challenges in building machine learning models can be tackled through systematic software engineering approaches.

Wan et al. ~\cite{8812912} studied how the adoption of machine learning changes software development practices. The work characterises the differences in various aspects of software engineering and task involved for machine learning system development and traditional software development. Lwakatare et al.
~\cite{10.1007/978-3-030-19034-7_14} propose a taxonomy that depicts maturity stages of use of machine learning components in the industrial software system and mapped the challenges to the machine learning pipeline stages.

Building machine learning models is becoming an engineering discipline where practitioners take advantage of tried-and-proven methods to address recurring problems~\cite{49406}.\\Washizaki et al.~\cite{8945075} studies the machine learning design patterns and architectural patterns. The authors also proposed an architectural pattern for machine learning for improving operational stability~\cite{8712157}. The work separates machine learning systems' components into business logic and machine learning components and focuses on the machine learning pipeline management, data management, and machine learning model versioning operations. 

The research on federated learning system design was first done by Bonawitz et. al~\cite{47976}, focusing on the high-level design of a basic federated learning system and its protocol definition. However, there is no study on the definition of architecture patterns or reusable solutions to address federated learning design challenges currently. Our work addresses this particular gap with respect to software architecture designs for federated learning as a distributed software system. To the best of our knowledge, this is the first comprehensive and systematic collection of federated learning architectural patterns. The outcomes are intended to provide architectural guidance for practitioners to better design and develop federated learning systems.

\section{Conclusion}
\label{S:conclusion}
Federated learning is a data privacy-preserving, distributed machine learning approach to fully utilise the data and resources on IoT and smart mobile devices. Being a distributed system with multiple components and different stakeholders, architectural challenges need to be solved before federated learning can be effectively adopted in the real-world. In this paper, we present 14 federated learning architectural patterns associated with the lifecycle of a model in federated learning. The pattern collection is provided as architectural guidance for architects to better design and develop federated learning systems. In our future work, we will explore the architectural designs that help improve the trust in federated learning.

\bibliographystyle{cas-model2-names}

\bibliography{Pattern_FL}


\end{document}